\documentclass[journal]{IEEEtran}
\usepackage{cite}

\ifCLASSINFOpdf
\else
\fi
\usepackage{amsmath}

\usepackage{graphicx}
\usepackage{amssymb}
\usepackage{booktabs}
\usepackage{multirow}
\usepackage{color}
\usepackage{bm}
\usepackage{threeparttable}

\usepackage[accsupp]{axessibility}  

\def\eg{\emph{e.g}.} 
\def\ie{\emph{i.e}.} 
 
\def\etc{\emph{etc}} \def\vs{\emph{vs}.}

\newcommand{\zlc}[1]{\textcolor{black}{#1}}
\newcommand{\zj}[1]{\textcolor{black}{#1}}
\newcommand{\zjn}[1]{\textcolor{black}{#1}}

\usepackage{footnote}
\makesavenoteenv{table}

\begin{document}
%
\title{Towards Explainable 3D Grounded Visual Question Answering: A New Benchmark and Strong Baseline}
%
%
%

\author{Lichen Zhao,
        Daigang Cai,
        Jing Zhang,
        Lu Sheng,
        Dong~Xu,~\IEEEmembership{Fellow,~IEEE}, \\
        Rui Zheng, Yinjie Zhao,
        Lipeng Wang and Xibo Fan
        \thanks{Lichen Zhao, Daigang Cai, Jing Zhang, Lu Sheng, Rui Zheng, Yinjie Zhao,
        Lipeng Wang and Xibo Fan are with the College of Software, Beihang University, China. E-mail: \{zlc1114, caidaigang, zhang\_jing, lsheng, rzheng, yjzhao, wanglipeng, fanxibo\}@buaa.edu.cn.}
        \thanks{Dong Xu is with the School of Electrical and Information Engineering, University of Sydney, Sydney, NSW, 2008 Australia. E-mail: \{dong.xu\}@sydney.edu.au.} \thanks{Corresponding author: Jing Zhang.}}

\maketitle

\begin{abstract}
Recently, 3D vision-and-language tasks have attracted increasing research interest. Compared to other vision-and-language tasks, the 3D visual question answering (VQA) task is less exploited and is more susceptible to language priors and co-reference ambiguity. 
Meanwhile, a couple of recently proposed 3D VQA datasets do not well support 3D VQA task due to their limited scale and annotation methods. 
In this work, we formally define and address a 3D grounded VQA task by collecting a new 3D VQA dataset, referred to as FE-3DGQA, with diverse and \zjn{relatively} free-form question-answer pairs, as well as dense and completely grounded bounding box annotations. 
\zjn{To achieve more explainable answers, we labeled the objects appeared in the complex QA pairs with different semantic types, including answer-grounded objects (both appeared and not appeared in the questions), and contextual objects for answer-grounded objects.}
We also propose a new 3D VQA framework to effectively predict the completely visually grounded and explainable answer. Extensive experiments verify that our newly collected benchmark datasets can be effectively used to evaluate various 3D VQA methods from different aspects and our newly proposed framework also achieves the state-of-the-art performance on the new benchmark dataset.
\end{abstract}


\section{Introduction}

In
recent years, there has been increasing research interest in various vision-and-language tasks~\cite{kazemzadeh2014referitgame,das2017visual_dialog,chen2015microsoft_coco_captions,das2018embodied_question_answering,anderson2018vision_and_language_navigation,iccv_AntolALMBZP15_VQAv1,cvpr_GoyalKSBP17_VQAv2,chen2020scanrefer,chen2021scan2cap}.
%
Among these tasks, visual question answering (VQA) systems aim to answer free-form questions based on the visual content in images or other types of data like 3D point cloud. 
\zlc{Compared to other vision-language tasks, the VQA task is more susceptible to language priors and co-reference ambiguity.
Without the visual data, it is not feasible to obtain the final results for other vision-language tasks.
For example, the captioning task merely takes visual data as the input and to generate textual descriptions. The grounding task aim to predict bounding boxes on the visual data.
And thus these tasks heavily rely on the understanding of the visual data. However, as shown in many previous works~\cite{iccv21_3D_Visual_Graph_Network_for_Object_Grounding,shih2016where_to_look} for the VQA task, even without the visual data, we can still achieve reasonable results by merely relying on textual questions, since some questions can be answered by common sense (\ie, language priors).
Thus the VQA task requires more consideration for both the design of the benchmark datasets and the corresponding methods.}

\begin{figure}[t]
\centering
\includegraphics[width=1\linewidth]{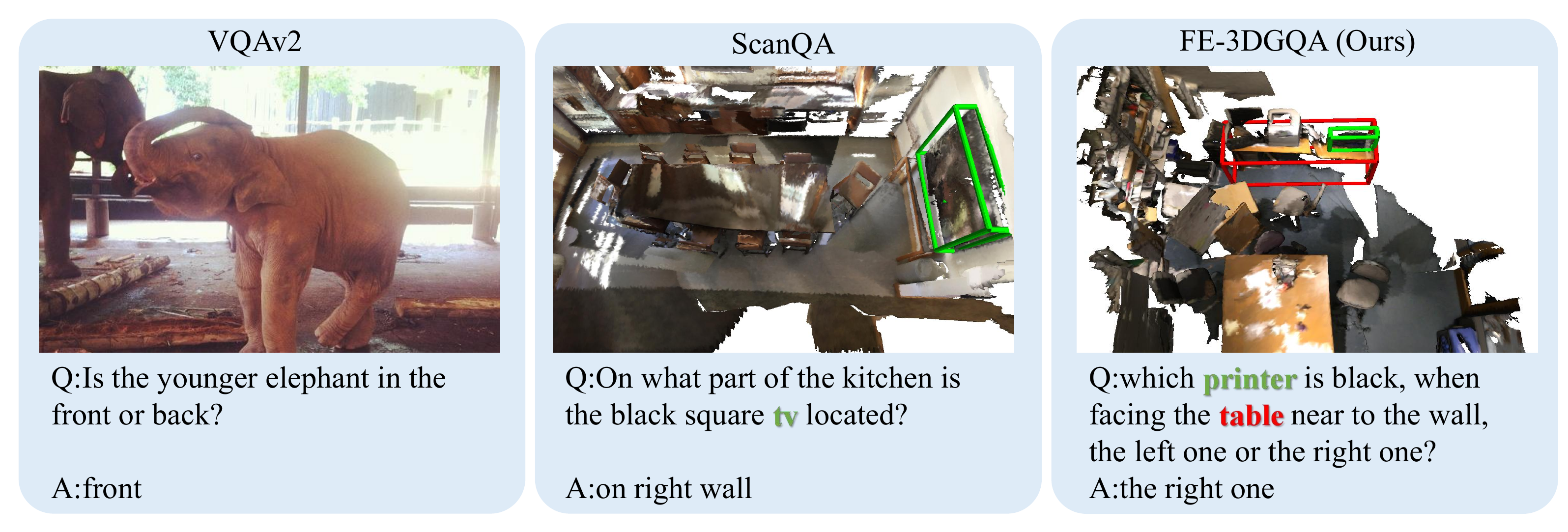}
\caption{Comparison of our collected dataset and other 2D VQA and 3D VQA datasets. When compared with ScanQA~\cite{corr2021_ScanQA}, our dataset contains complex relations with finely-annotated 3 types of related objects. When compared to 2D data, the 3D VQA task can avoid the inherent spatial ambiguity (\eg, ``front'').}
\label{fig:2d_comparison}
\vspace{-4mm}
\end{figure}

Compared to the 2D VQA tasks, the VQA task based on the 3D visual scenes is less exploited, which introduces more opportunities and challenges to the 3D vision-and-language field. \zlc{As shown in Fig.~\ref{fig:2d_comparison}, extending the 2D VQA task to 3D brings more accurate relative directions and distances in the original 3D scenes, which cannot be well preserved in 2D images.} On one hand, compared to the 2D images, the 3D point cloud-based visual scenes contain rich, complex, and less biased geometric and relation information of different objects without the issues of occlusion.
\zlc{Thus, more \emph{complex and relatively free-form questions} can potentially be correctly answered with the aid of the rich 3D point cloud data. On the other hand, to correctly answer these complicated questions, more information related to the objects appeared in the question should be involved to obtain \emph{reliable and explainable results}}. For example, not only the answer-grounded objects should be grounded by the model, the contextual objects related to the answer-grounded objects should also be identified to precisely predict the correct answer to the questions that involve complex object relationships. Therefore, a \zjn{\emph{complex and explainable}} 3D grounded visual question answering task needs to be introduced and formulated to boost the development of the 3D visual-and-language field.

To tackle the 3D VQA task, the model needs to first ground to the related objects and then identify how they are related to the answers in the 3D scene based on the textual question, and then infer the correct answer based on the question and the grounded objects. In this way, the model can achieve the explainable 3D VQA results.
In this work, we formulate and address the relatively free-form and explainable 3D VQA task by defining a quintuple (\ie, $\langle$question, 3D point cloud data, answer, completely grounded object labels, completely grounded object location$\rangle$) for the 3D VQA model. The inputs to the model include a 3D point cloud together with a complex question based on the 3D scene. The outputs from the model contain not only the correct answer to the question, but also the class labels and localization of the related objects (as well as how they are related to the answer) in the scene to assist the reasoning and evaluation of the correct answer. 
It is worth mentioning that more complex relationships could be involved for the 3D VQA task when compared to the 2D VQA task. Thanks to the rich geometric and relation information contained in 3D point cloud data,
we formally and fine-grainedly define three types of objects required to reliably assist the inference and evaluation of the correct answer to the questions  in 3D VQA: 1) AG-In-Q: the answer-grounded objects appeared in the question, 2) AG-NotIn-Q: the answer-grounded objects not appeared in the question, and 3) Context-Of-AG: the contextual objects related to the answer-grounded objects appeared in the question. Given the question and the 3D scene, the three types of objects jointly provide the reasoning evidence and explanation to the predicted answers. Thus, if all the three types of objects are grounded (including their class labels and localizations as well as their correct object types), we call the 3D VQA task would be completely grounded and explainable.

\begin{figure}[t]
\centering
\includegraphics[width=1\linewidth]{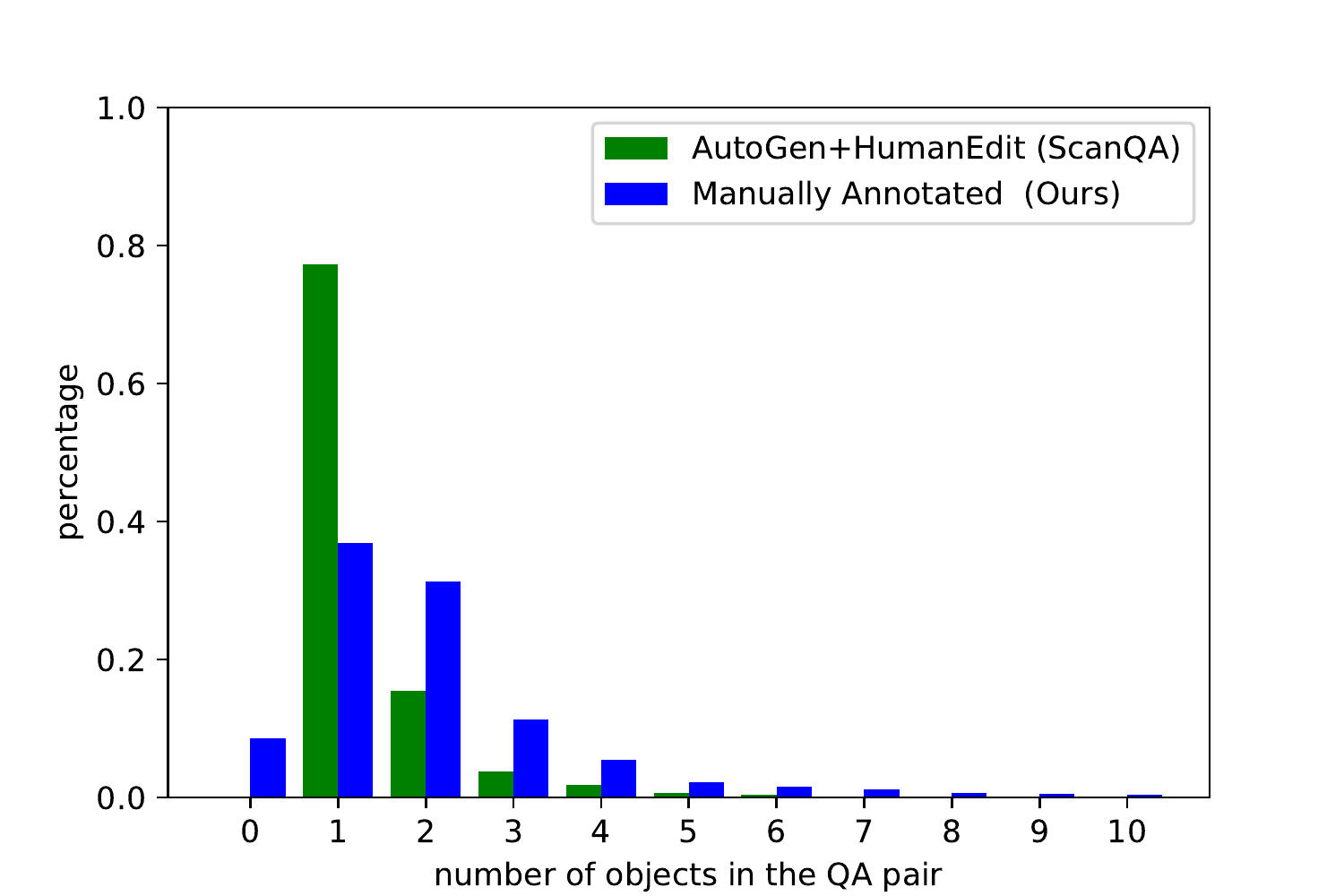}
\caption{The distribution of the number of annotated QA-related objects of our FE-3DGQA dataset and existing AutoGen+HumanEdit dataset ScanQA~\cite{corr2021_ScanQA}. \zlc{Our proposed dataset includes complete object annotations for different types of objects in each QA, providing explainability for 3D VQA tasks.}} 
\label{fig:rebuttal_distribution}
\end{figure}

Though three 3D VQA datasets~\cite{corr2021_3D_Question_Answering,corr2021_CLEVR3D,corr2021_ScanQA} are proposed recently, none of these datasets could well support the free-form and explainable 3D grounded VQA task due to their limited annotations of the question-answer pairs and incomplete annotations to the grounded objects (see \zlc{Fig.~\ref{fig:rebuttal_distribution} and} Sec.~\ref{sec:related_work_3d} for more details).
To this end, in this work, we collect a new 3D visual question answering dataset referred to as FE-3DGQA with diverse and complex question-answer pairs together with dense and complete grounded object bounding box annotations. Based on the ScanNet dataset~\cite{dai2017scannet}, our dataset consists of 20k manually annotated question-answer pairs with an average of 2.1 completely grounded objects' bounding box annotations for each QA pair from 703 indoor scenes. Different from the previous works~\cite{corr2021_CLEVR3D,corr2021_3D_Question_Answering,corr2021_ScanQA} without or with only one type of grounded objects, 
we annotate the aforementioned three different types of grounded objects for each QA pair. 
In addition, observing that all the existing 3D VQA datasets are limited in scale when compared with their 2D counterparts, we further expand our FE-3DGQA dataset by using the existing ScanNet~\cite{dai2017scannet} (\eg, ScanRefer~\cite{chen2020scanrefer} and Referit3D~\cite{eccv2020_referit3d}) datasets based on question templates, resulting in an Extended FE-3DGQA dataset with 112k additional question-answer pairs. 

Building upon the defined free-form and explainable 3D VQA task and the two newly collected FE-3DGQA datasets, we for the first time propose a strong baseline.
\zlc{Even though much progress has been made in VQA on mult-view 2D images from 3D scenes, it is non-trivial to fuse the results from different views as these results may be inconsistent due to the missing relative directions and occlusions in images.}
\zlc{Existing 3D VQA works~\cite{corr2021_ScanQA,corr2021_CLEVR3D,corr2021_3D_Question_Answering} encode and fuse the questions and visual data by using different methods, such as Bi-LSTM, BERT, and other encoder-decoder networks. In contrast, we explicitly enhance the within-object attribute features and model the across-object relations by using an enhanced self-attention module, and carefully design our framework to effectively explore three types of related-objects.}
Specifically, our method tackles the problem by three objectives: 1) grounding and attending to all the related objects in the 3D scene based on the question, 2) identifying how these objects are related to the answer, and 3) inferring the correct answer based on the question and different types of grounded objects. To achieve the three objectives, we design an end-to-end optimized 3D VQA method with a language branch, a 3D vision branch, and a fusion module. 
The language branch is based on a pre-trained T5 model to handle the free-form question-answer pairs. 
It provides the object labels of ``AG-In-Q'' and ``Context-Of-AG'' objects to guide the corresponding attentions in the 3D scene. 
Moreover, it also limits the scope of the candidate answers and links to the potential answers of the question. The 3D vision branch includes a 3D object detector to tokenize the 3D scene with the object proposals as the object tokens, followed by a token enhancement module to extensively encode the within-object and across-object features of all the related objects, which can effectively help additional identify the ``AG-NotIn-Q'' objects,
and thus ground and attend to all three different types of related objects and eventually provide the visual evidence to the answer. The co-attention-based fusion module enables the interactions between the question and 3D point clouds, which establishes various semantic links between textual question-answer pairs and the corresponding object regions to filter out unrelated objects (and attend to the related objects) in the 3D point clouds, and thus allows the correct predictions of the final answer based on the joint textual and visual features in the subsequent grounded answer generation module.

The contributions of our work can be summarized as: (1) We define a new 3D vision and language task called free-form and explainable grounded 3D VQA. To this task, we also collect the new dataset called FE-3DGQA with 20k free-form question-answer pairs, in which we additionally annotate three types of corresponding grounded objects from 703 3D scenes. (2) A new framework is proposed, which can be used as a strong baseline for the free-form and explainable grounded 3D VQA task. (3) The proposed framework is extensively evaluated on the proposed FE-3DGQA dataset and achieves the state-of-the-art results.

\section{Related Work}

\subsection{Vision And Language on 3D scenes.}
Deep learning in various 3D point cloud based vision tasks has attracted a great deal of interest~\cite{guo2021JointPruning,wangkai2021sequential,liu2021geometrymotion,csvt_zhao2021transformer3d,cvpr_QiSMG17_PointNet,tcsvt_SongZZ22,tcsvt_WangZLM21,tcsvt_MekuriaBC17,tcsvt_SongSGWL21}.
Unlike 2D datasets, the data collection with 3D annotations are expensive, which limits the development of the unified 3D framework. Some recent works are also proposed to explore the 3D vision+language tasks for scene understanding, such as visual grounding~\cite{chen2020scanrefer,eccv2020_referit3d,iccv21_3DVG-Transformer}, visual captioning~\cite{chen2021scan2cap,arXiv_2021d3net}, 3D scene graph~\cite{cvpr_Wald2020_3DSSG}, \etc.
ScanRefer~\cite{chen2020scanrefer} and Referit3D~\cite{eccv2020_referit3d} are two datasets to first introduce the task to localize 3D objects on RGB-D indoor Scans. Based on the ScanRefer~\cite{chen2020scanrefer} dataset, Scan2Cap~\cite{chen2021scan2cap} is recently proposed for dense captioning on 3D scenes, which focuses on describing the attribute and relation information of objects in the scene. 3DSSG~\cite{cvpr_Wald2020_3DSSG} is proposed for 3D scene graph generation, which is built based on several instances from the 3RScan~\cite{iccv_Wald2019RIO_3RScan} dataset. 
%
Different methods~\cite{chen2020scanrefer,eccv2020_referit3d,yuan2021instancerefer,chen2021scan2cap,aaai_huang2021_TGNN,iccv21_3DVG-Transformer,PMLR21_corl_RohDFF21_LanguageRefer,iccv21_sat,iccv21_3D_Visual_Graph_Network_for_Object_Grounding,cvpr21_sunrefer,he2021_transrefer3d,wacv_AbdelreheemUSYC22_3DRefTransformer,arXiv_2021d3net,corr2021_ScanQA,corr2021_CLEVR3D,corr2021_3D_Question_Answering,cvpr_Wald2020_3DSSG} are proposed to tackle these problems by joint modeling the language and vision data.
However, compared to these 3D vision-and-language tasks, the 3D visual question answering is more prone to ignore the visual data and overfit to the language priors. Thus, both the collection of 3D VQA dataset and design of 3D VQA method require more consideration compared to other 3D vision-and-language tasks.

\subsection{Visual Question Answering (VQA).}
\label{sec:related_work_3d}
Several 2D image-based VQA datasets~\cite{nips_MalinowskiF14_DAQUAR,iccv_AntolALMBZP15_VQAv1,cvpr_GoyalKSBP17_VQAv2} and methods~\cite{shih2016where_to_look,zhou2015simple_baseline,teney2018tips_and_tricks,lu2016hierarchical,li2020oscar,lu2019vilbert} are proposed in recent years. However, the VQA task based on 3D point clouds-based visual scene is far less exploited, which introduces more opportunities and challenges to the 3D vision-and-language field.
Recently, three datasets~\cite{corr2021_3D_Question_Answering,corr2021_ScanQA,corr2021_CLEVR3D} were collected for 3D visual question answering. For example, 3DQA~\cite{corr2021_3D_Question_Answering} only contains 6k manually annotated questions, and they do not provide the related object labels and bounding boxes. CLEVR3D~\cite{corr2021_CLEVR3D} is built upon the 3D scene graph datasets and their question-answer pairs are created based on a set of fixed templates from the scene graph annotations. In other words, their annotations are not fully free-form and highly rely on the scene-graph annotations, and thus cannot be easily expanded to other 3D scenes without scene-graph annotations. ScanQA~\cite{corr2021_ScanQA} is another 3D VQA dataset, which contains 41k question-answer pairs. However, their questions are automatically generated from the question generation model based on the fixed form grounding annotations. Though they have annotated the corresponding 3D bounding boxes related to the answers, the annotations of objects are not elaborately classified by how they are related to the questions or answers, 
which limits their full reasoning capability on how the answers are generated by the VQA models. 

In summary, none of the 3D VQA datasets could well support the free-form and explainable 3D grounded VQA task due to their limited annotation methods of the question-answer pairs and incomplete annotations to the grounded objects.
In contrast, we collect a new FE-3DGQA dataset, which is a well-annotated dataset with various annotations for object-of-interest for free-form and explainable QA.
We argue that our dataset could help evaluate language-based 3D scene understanding methods, and help reduce the bias issue from the language prior. Moreover, we design a new completely grounded and explainable 3D VQA framework by using a joint training strategy with a specifically designed language branch, an enhanced 3D vision branch, and a fusion module for both VQA and visual grounding tasks. 

\section{FE-3DGQA Dataset}

\begin{figure}[t]
\centering
\includegraphics[width=\linewidth]{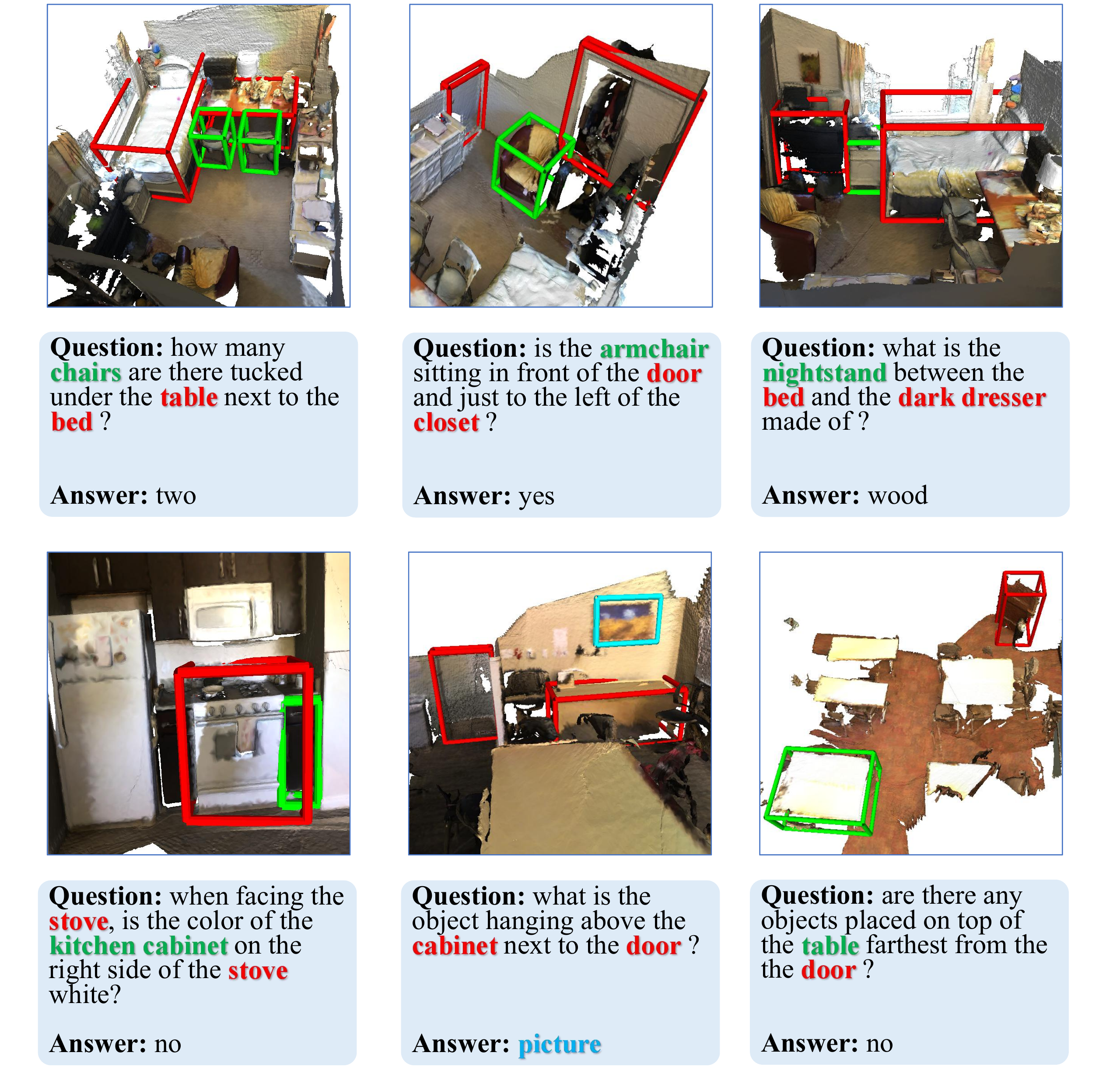}
\caption{Visualizations of the newly collected FE-3DGQA dataset. The related objects of AG-In-Q, AG-NotIn-Q, and Context-Of-AG are colored in green, cyan, and red, respectively.}
\label{fig:visualization_gt}
\end{figure}


Compared to other vision-language tasks, the VQA task is more susceptible to language priors and co-reference ambiguity and thus requires more consideration for both the design of the benchmark datasets and the corresponding methods. For example, if the questions and answers are highly correlated, the models tend to make predictions based on the co-occurrence patterns between textual questions and answers (\ie, the language priors) instead of reasoning upon visual contents. In addition, if the question-answer pairs are not designed properly, the co-reference ambiguity issue arises when a referred object in the questions or answers has multiple correspondences in an image, and the context of the questions or answers is insufficient to distinguish them.
When compared to the 2D VQA task, the VQA task based on the 3D scenes introduces more opportunities and challenges, which motivates us to collect our FE-3DGQA dataset. Fig.~\ref{fig:visualization_gt} shows some representative examples from our FE-3DGQA dataset. Each 3D VQA sample contains a question based on a 3D scene, the corresponding answer, and the completely grounded object annotations in the 3D scene to facilitate the subsequent research to solve the relatively free-form and explainable 3D grounded VQA task. 
Our FE-3DGQA dataset consists of two parts: 1) the relatively free-form question-answer pairs, 2) the completely grounded object annotations. 

\subsection{FE-3DGQA Dataset Collection}
\label{sub:3d_vqa_dataset_collection}
Our dataset is annotated based on ScanNet~\cite{dai2017scannet}, which contains the RGB-D Scans of various indoor scenes with fine-grained instance-level segmentation annotations.

\noindent\textbf{Complex and Relatively Free-Form Question-Answer Pairs.}
Different from the 2D image, the 3D point cloud-based visual scenes contain rich, complex, and less biased geometric and related information of different objects without the occlusion issue. Thus, more complex \emph{free-form} questions can potentially be correctly answered with the aid of the rich 3D point cloud data. As shown in Fig.~\ref{fig:visualization_gt}, our FE-3DGQA dataset contains more complex and free-form questions, where full exploitation and understanding of the corresponding 3D scene is often required to answer those questions. 

Unlike the previous template-based or auto-generated question-answer pairs based on the existing ScanRefer~\cite{chen2020scanrefer} dataset, our datasets are manually annotated with proper instructions on the design of non-trivial questions. Thus, our annotations are completely free-form and potentially go beyond the ScanRefer grounding dataset and ScanNet detection datasets with more complex multi-object relationships. In addition, our annotators are instructed to carefully design the questions customized to each individual scene in a whole to maximally avoid co-reference ambiguity with respect to both questions and answers. 
Specifically, to encourage the diversity of QA pairs to reflect a broad coverage of 3D scene details, we require the annotators to annotate each object in a scene with at least 2 different questions. We do not provide any question template to avoid limiting the richness of the questions. However, the questions in each scene should cover all the four different aspects in free-form: 1) the local object-oriented questions with attributes and local relations of a specific object (\eg, ``what color is the table behind the l-shaped couch?'', ``what shape is the table between two white chairs?'', ``what is between two printers on the desk?''), 2) the global context-aware questions to understand the whole scene (\eg, ``what is this room?'', ``how many printers in the room?''), 3) the complex relationships among multiple objects (\eg, ``which is closer to the curtain, the suitcase or the nightstand?''), and 4) the direction or location related questions potentially for navigation (\eg, ``where is the couch?'', ``what kind of furniture is facing two green chairs?'').

\noindent\textbf{Completely Grounded Objects Annotations.} To support and enable the development of fully explainable 3D VQA models, we not only design and annotate the question-answer pairs, but also annotate all the grounded objects related to the question-answer pairs. Though the ScanQA~\cite{corr2021_ScanQA} dataset also provides the annotations of the answer-grounded objects, we take one step further, and fully annotate all the objects (with object classes and locations) related to both questions and answers. 
Moreover, we fine-grainedly categorize the grounded objects into the following three types:
\begin{enumerate}
    \item \emph{AG-In-Q: the answer-grounded objects appeared in the question} (\eg, the `table' in the question `what color is the table behind the l-shaped couch?').
    \item \emph{AG-NotIn-Q: the answer-grounded objects not appeared in the question} (\eg, the `chair' in the question `what is under the white table behind the l-shaped couch?' and the answer `chair').
    \item \emph{Context-Of-AG: the contextual objects related to the answer-grounded objects appeared in the question} (\eg, the `l-shaped couch' in the question `what color is the table behind the l-shaped couch?').
\end{enumerate}
The three types of objects jointly assist the design and evaluation of the explainable 3D VQA methods. To be specific, the ``AG-In-Q'' and ``AG-NotIn-Q'' objects are answer-grounded. As long as these two types of objects are correctly grounded, the answer to the question will be more explainable. Otherwise, the question may be obtained merely from the language priors in the dataset or by the common sense based on the questions. Moreover, the ``AG-In-Q'' and ``AG-NotIn-Q'' annotations also enable the visual answers in the form of bounding boxes. In addition, the ``Context-Of-AG'' objects appeared in the questions are related to the answer-grounded objects, which are also the key to predict the correct answer. None of the existing 3D VQA datasets annotate this type of objects. However, we argue that the grounding results of the ``Context-Of-AG'' objects in the visual data provide the key reasoning evidence of the correct answer. For example, for the question: ``what is between the two printers on the desk?'', both of the ``two printers'' and the ``desk'' belong to the ``Context-Of-AG'' objects. Thus, the ``Context-Of-AG'' objects provide the rich cues to the correct answer. So the model needs to precisely ground to the ``Context-Of-AG'' objects before correctly answering the question.
To aid the prediction and explanation of the correct answer, we annotate the bounding boxes of three different types of objects related to the answers as shown in Fig.~\ref{fig:visualization_gt}. The ``AG-In-Q'' objects are marked in green, which indicates the answer-grounded objects that appeared in the question. The ``AG-NotIn-Q'' objects marked in cyan are answer-grounded objects not appeared in the question. The contextual objects of the answer-grounded objects are categorized as ``Context-Of-AG'', which are marked in red. To precisely answer the complex questions with a reasonable explanations, all the three potential types of objects should be correctly grounded. 

\begin{table*}[t]
\begin{center}
\caption{Comparison between our dataset and other 3D VQA datasets.}
\vspace{-1em}
\begin{tabular}{c|c|c|c|c}
\hline
\textbf{Datasets} & \textbf{Source} & \textbf{Amount} & \textbf{Object Annotations} & \textbf{Collection} \\ \hline
ScanQA~\cite{corr2021_ScanQA} & ScanNet~\cite{dai2017scannet} & 41k & Unknown Types & AutoGen+HumanEdit \\
ScanQA~\cite{corr2021_3D_Question_Answering} & ScanNet~\cite{dai2017scannet} & 6k & Not annotated & Human \\
CLEVR3D~\cite{corr2021_CLEVR3D} & 3RScan~\cite{iccv_Wald2019RIO_3RScan} & 60k & Not annotated & Template \\ \hline
FE-3DGQA (Ours) & ScanNet~\cite{dai2017scannet} & 20k & AG In / NotIn Q \& Context of AG & Human \\
Ext. FE-3DGQA (Ours) & ScanNet~\cite{dai2017scannet} & 112k & AG In / NotIn Q & Template \\ \hline
\end{tabular}
\label{tab:comparison_with_other_datasets}
\end{center}
\end{table*}

\subsection{FE-3DGQA Dataset Statistics.}

%
We collected $20,215$ question-answer pairs for $703$ ScanNet~\cite{dai2017scannet} scenes
\footnote{After excluding 97 test set scenes (due to Copyright issues) from the total 800 scenes, we follow ScanRefer[6] to only annotate one single scan for each of 703 scenes.}
with $42,456$ annotated grounded objects. The average numbers of three different types of object in each 3D scene are $1.53$, $0.18$, and $0.39$, respectively. Overall, we annotated $2.1$ objects on average for each scene. And we follow the official ScanNet~\cite{dai2017scannet} setting to split FE-3DGQA as the train/val set, with $16,245$ and $3,970$ samples, respectively. On average, there are $9.10$ words per question and $28.76$ question-answer pairs per scene. 
Our manually labeled FE-3DGQA dataset enjoys the advantage of free-form and completely grounded objects. However, it is still limited by relatively small scale when compared to other 2D VQA datasets. To enrich our FE-3DGQA dataset, we further extend our 3D VQA dataset by including the original ScanNet~\cite{dai2017scannet} dataset, ScanRefer~\cite{chen2020scanrefer} dataset, and Referit3D~\cite{eccv2020_referit3d} dataset into their QA-like versions based on the fixed templates. We name the extended FE-3DGQA dataset as Ext. FE-3DGQA, which contains around 112k additional template-based QA pairs and 100k additional grounded objects. 


Table~\ref{tab:comparison_with_other_datasets} shows the comparison between the existing concurrent 3D VQA datasets~\cite{corr2021_3D_Question_Answering,corr2021_CLEVR3D,corr2021_ScanQA} and our newly collected FE-3DGQA and the Ext. FE-3DGQA datasets. It can be seen that our FE-3DGQA dataset is the largest manually annotated 3D VQA dataset, with complete grounded object annotations. Moreover, by using additional template-based QAs, our Ext. FE-3DGQA dataset is the largest 3D VQA dataset.

\begin{figure}[t]
\centering
\includegraphics[width=\linewidth]{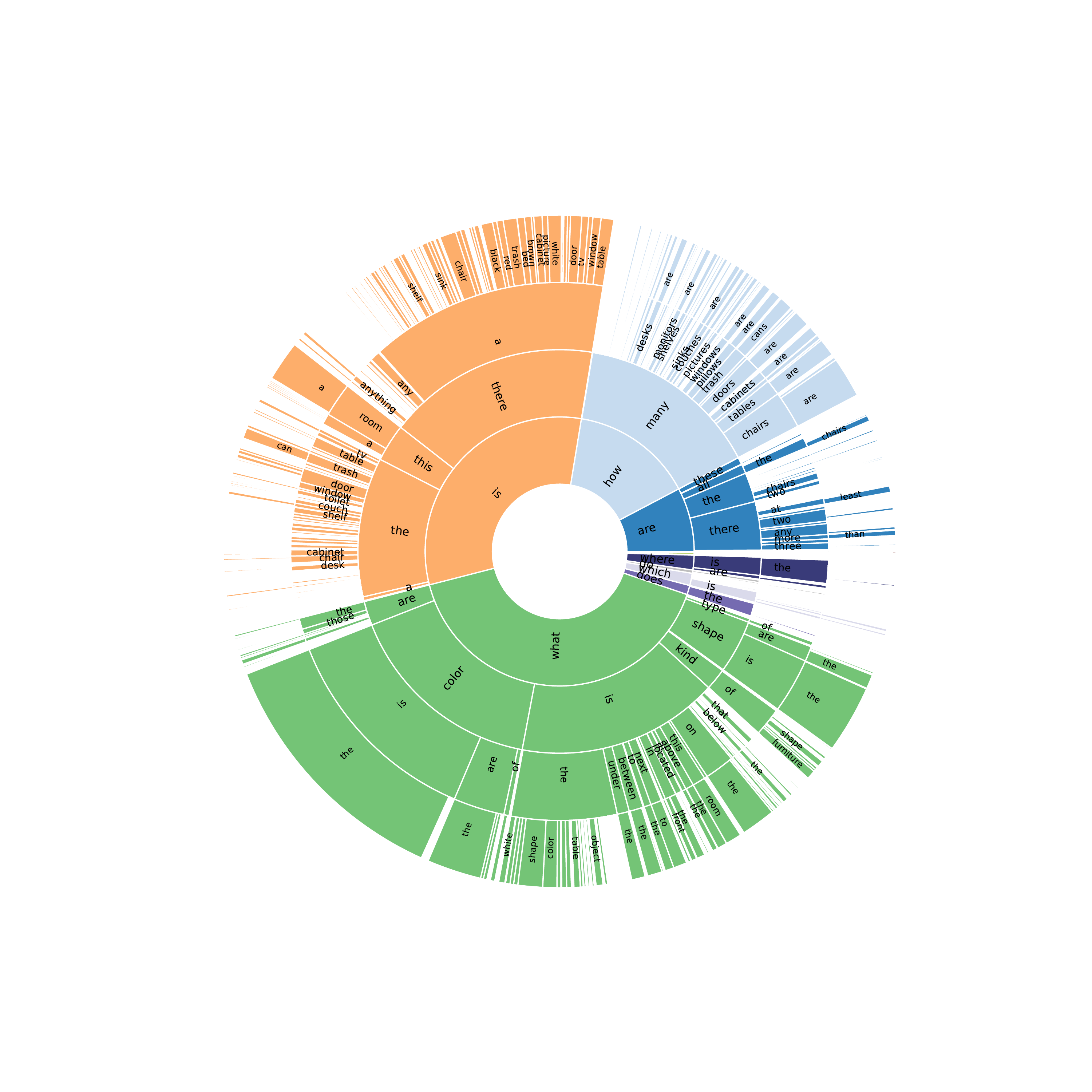}
\caption{Question distribution in terms of their first four words in FE-3DGQA.}
\label{fig:question_distribution}
\end{figure}

\begin{figure}[t]
\centering
\includegraphics[width=\linewidth]{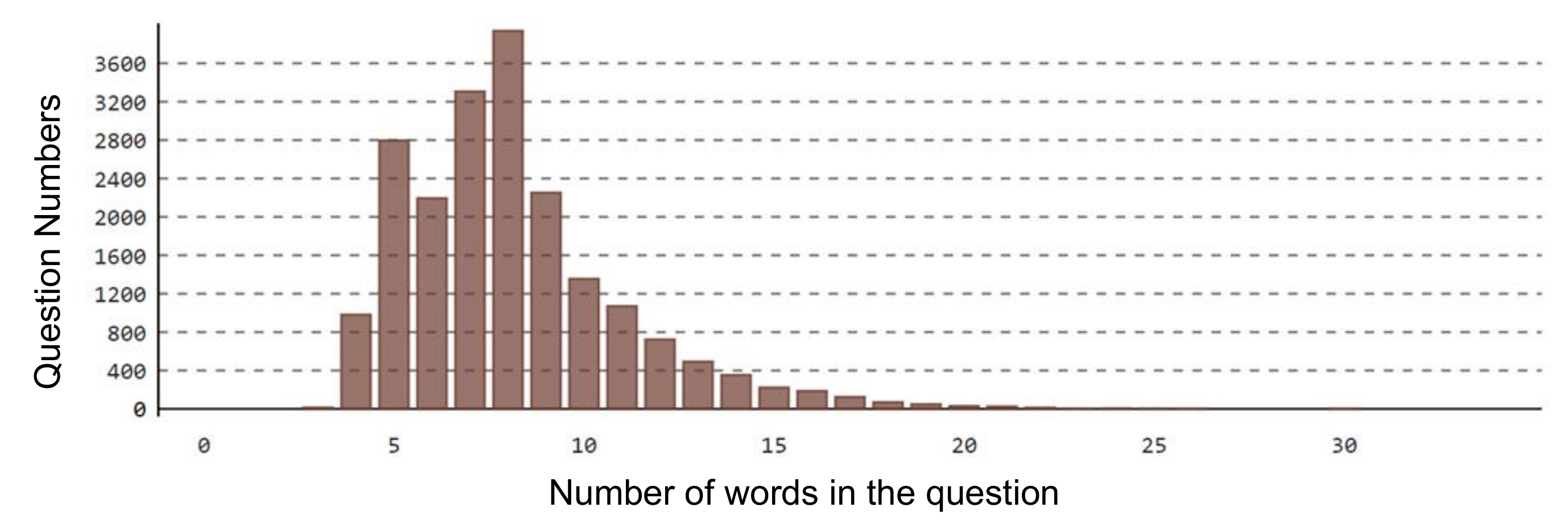}
\caption{Question distribution in terms of the number of words in each question from FE-3DGQA.}
\label{fig:question_length_distribution}
\end{figure}

\begin{figure*}[t]
\centering
\includegraphics[width=0.96\linewidth]{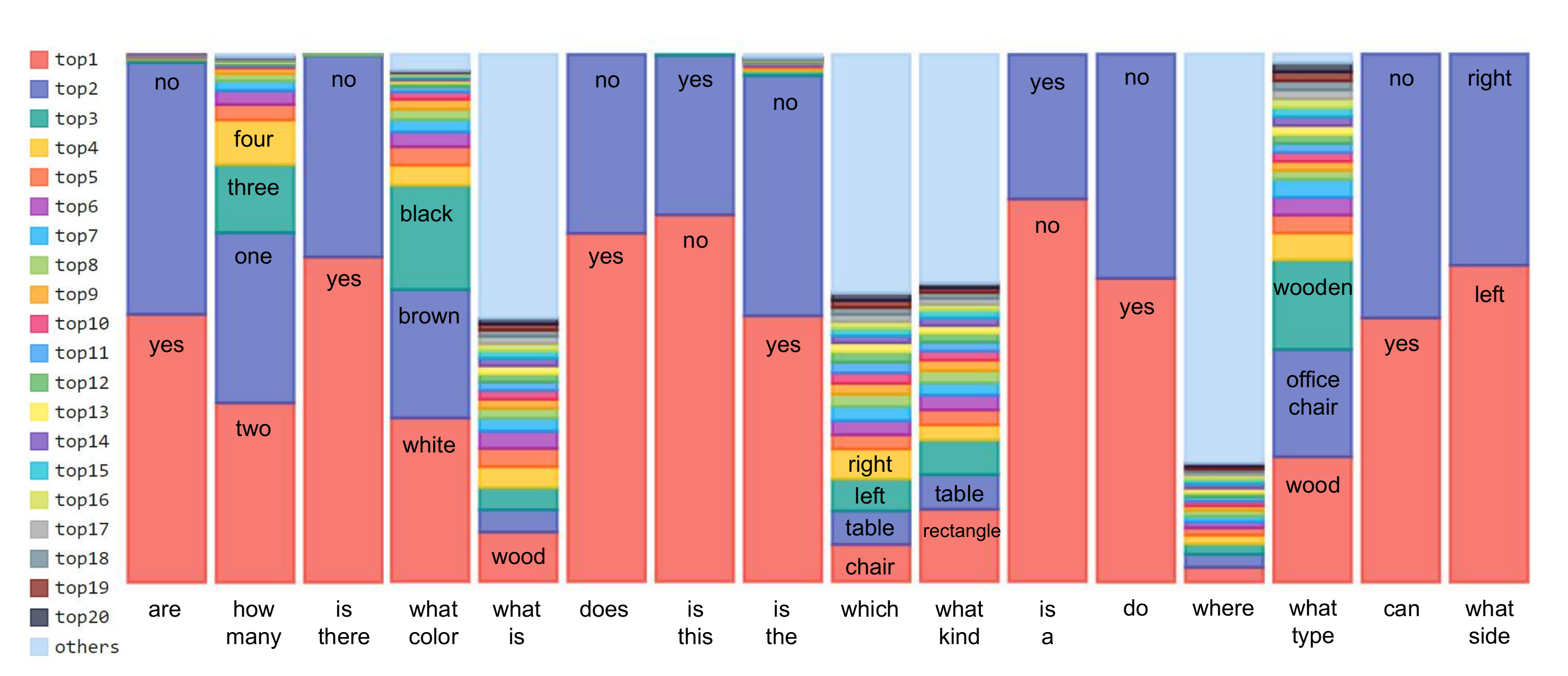}
\caption{Answer distribution in terms of different type of questions in FE-3DGQA.}
\label{fig:answer_distribution}
\end{figure*}




As shown in Fig.~\ref{fig:question_distribution}, we report the distribution of the question types based on the first four words of the questions in FE-3DGQA. It can be seen that all four aspects are involved in the questions. For example, 1) the local object-oriented questions (\eg, `what color is the', `what kind of furniture'), 2) the global context-aware questions (\eg, `what is this room?', `are there more than', `how many chairs are'), 3) the complex relationships among multiple objects (\eg, `which is more/less/taller', `what is between the'), and 4) the direction or location related questions potentially for navigation (\eg, `where is the', `facing the').

In Fig.~\ref{fig:question_length_distribution}, we show the distribution of the question types based on the number of words in each type. We can see that the questions have varied lengths with 9.10 words on average, which further demonstrates the diversity of our questions. 
Fig.~\ref{fig:answer_distribution} shows the distribution of answers 
\zj{to the question types by the first few words with the highest numbers of samples. We can observe that the answers to most of the questions are relatively balanced.}




\subsection{FE-3DGQA dataset collection details}




\noindent\textbf{Annotation.} 
To ensure the quality and diversity of the grounded questions, especially the correctness of bounding box annotations of various related objects, we ask well-educated university students to annotate the scenes based on an online visualizer website, on which annotators could interactively rotate, move, zoom in/out, and obtain object names and ids. We also developed the scripts to visualize different types of related objects to check the annotations offline. 
Specifically, the students are divided into six groups, and each group is assigned with a well-trained leader annotator to ensure the quality of the annotations. 
Furthermore, we provide more than 200 samples based on 12 different 3D point cloud scenes to provide concrete instructions for the annotators.
Each object in a scene is annotated with at least two different questions, and each scene should cover all four aspects to achieve the diversity of QA pairs and reflect a broad coverage of 3D scene details. In addition, our annotators are instructed to carefully design the questions customized to each individual scene in a whole to maximally avoid co-reference ambiguity with respect to both questions and answers.
To avoid the view dependency issue in the question (\ie, spatial relations such as ``left'' in the 3D-scene is unclear when facing different directions), we inspire the annotators to add the detailed direction constraints when there are complex relationships in the question. 

\noindent\textbf{Filtering.}
After collecting the QA pairs together with the objects, each question is refined and filtered more than twice by two different additional annotators.
A different annotator first checks and filters the annotated questions and objects, and the group leader will double-check all the annotations again.
In the filtering stage, the question, the answer, and the related objects are checked and filtered simultaneously. 
\section{Extended FE-3DGQA Dataset}

\begin{table*}[t]
\caption{Data statistics of our FE-3DGQA dataset and the extended FE-3DGQA dataset.
The numbers in the columns `AG-In-Q', `AG-NotIn-Q', and `Context-Of-AG' indicate the average number of `the answer-referred objects appeared in the question', `the answer-referred objects not appeared in the question', and `the contextual objects related to the answer-referred objects appeared in the question' in each question, respectively. The numbers in the column ``Overall'' indicates the average number of all types of related object annotations in each question. 
${[+]}$: Transformed by the extending method described in Section~\ref{sec:Ext}. 
${[*]}$: 
Since the transformed grounding dataset and the transformed masked grounding dataset are based on the same dataset, we count the total number of QA pairs and the total number of related objects transformed from the same dataset only once in the Ext. FE-3DGQA dataset.
${[-]}$: Not annotated.
}
\centering
\begin{tabular}{c|cc|cccc}
\hline
Datasets & \#QA & \#Related Objects & AG-In-Q & AG-NotIn-Q & Context-Of-AG & Overall \\ \hline
FE-3DGQA & 20,215 & 42,456 & 1.530 & 0.178 & 0.392 & 2.100 \\ \hline
ScanNet$^{+}$ & 24,539 & 12,930 & 0.527 & 0 & 0 & 0.527 \\
ScanRefer$^{+}$ & 46,173 & 46,173 & 1 & - & - & 1 \\
Nr3D$^{+}$ & 41,475 & 41,475 & 1 & - & - & 1 \\
Masked ScanRefer$^{+}$ & 46,173 & 46,173 & - & 1 & - & 1 \\
Masked Nr3D$^{+}$ & 41,475 & 41,475 & - & 1 & - & 1 \\ \hline
Ext. FE-3DGQA & 112,187$^{*}$ & 100,578$^{*}$ & 0.503 & 0.439 & 0 & 0.942 \\ \hline
\end{tabular}
\label{tab:ext_statistics}
\end{table*}

\subsection{Extending existing datasets to GQA-like version}
\label{sec:Ext}
To enrich our FE-3DGQA dataset, we further extend our 3D VQA dataset by transforming the original ScanNet~\cite{dai2017scannet} dataset, the ScanRefer~\cite{chen2020scanrefer} dataset, and the Referit3D~\cite{eccv2020_referit3d} dataset into their QA-like versions based on the fixed templates for better scene understanding.
The statistics of the extended FE-3DGQA dataset are shown in Tab.~\ref{tab:ext_statistics}.
Below we describe how we transform each dataset into our Ext. FE-3DGQA in detail.

\noindent\textbf{Transforming 3D Detection Dataset to 3D GQA.}
The detection dataset consists of the dense object class annotations, which can be transformed to some reasonable questions with fixed templates. Here, we use the same detection annotations from the ScanRefer dataset, in which the class names are slightly different from the original ScanNet~\cite{dai2017scannet}.
To be specific, we use some templates to generate a new QA dataset based on the annotations from the original detection dataset (\eg, ``how many $<object>$ are there in the room?'', or ``is there a $<object>$ in the room?''). The blanks in the template are filled with the 17 candidate object classes contained in the ScanNet dataset. In this way, all the referred objects are marked as ``AG-In-Q'' objects. Considering that for many answers, we have ``zero'' object to be referred to in this extended dataset, we eventually obtain 24,539 QAs with an average of 0.527 annotated objects for each query in the scenes.

\noindent\textbf{Transforming 3D Visual Grounding Datasets to 3D GQA.}
Except for the detection annotations, the ScanRefer dataset and the Nr3D dataset for the visual grounding task also provide rich grounding annotations with dense descriptions of object attributes and relationships.
Thus, we treat the descriptions from the object grounding task as the input question, the referred object name as the answer, and the corresponding object bounding box annotation as the visually grounded ``AG-In-Q'' object.

\noindent\textbf{Transforming Masked 3D Visual Grounding Datasets to 3D GQA.}
Since the answers (\eg, the grounded objects) in the transformed visual grounding datasets can always be found in the questions (\eg, the grounding description), the VQA model may be biased by the language priors. To alleviate the issue, we propose to mask the referred object in the grounding description to form a new question.
We treat the masked object name as the answer and the corresponding object bounding box annotation as the visually grounded ``AG-NotIn-Q'' object.

\section{Methodology}

\begin{figure*}[t]
\centering
\includegraphics[width=\linewidth]{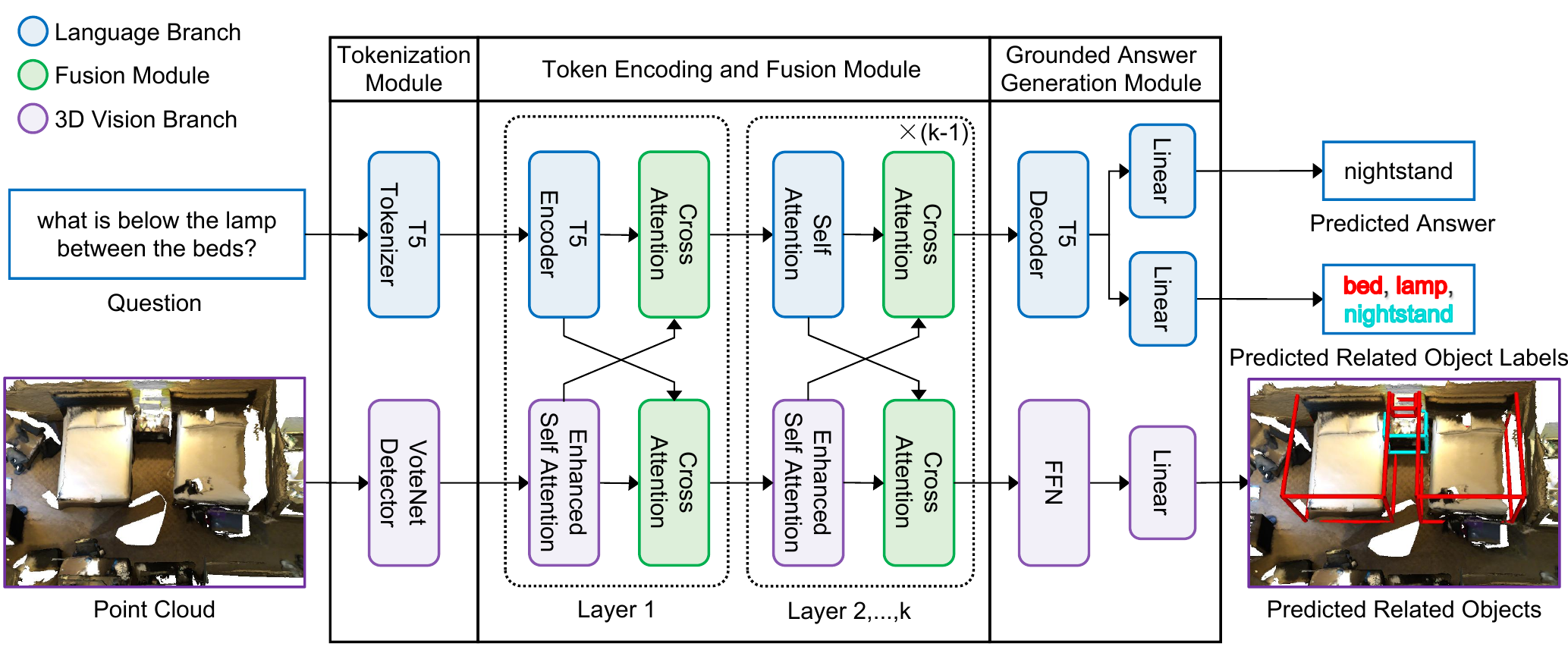}
\caption{Overview of our framework. The inputs (\ie, the question and the point cloud) are first encoded by the tokenization module. Then, the word tokens and the object tokens are enhanced and fused by using several transformer-based layers, which consist of both self-attention and co-attention modules. Finally, the grounded answers together with the predicted bounding boxes as well as their types and labels are generated by using separate decoders.}
\label{fig:pipeline}
\end{figure*}

With our completely grounded 3D VQA dataset, we formulate the FE-3DGQA problem as follows: given the point cloud $\boldsymbol{P}$ and the question $\boldsymbol{Q}$ about the 3D scene, we introduce a new framework to predict the visually grounded answer together with the corresponding bounding boxes and labels of all the related objects.
Our method tackles the problem by three objectives: 1) grounding and attending to all the related objects in the 3D scene based on the question, 2) identifying how these objects are related to the answer, and 3) inferring the correct answer based on the question and different types of grounded objects. To achieve these objectives, as shown in Fig.~\ref{fig:pipeline}, our framework consists of a language branch, a 3D vison branch, and a fusion module. We further split our model into the tokenization module to tokenize both the 3D point clouds and the textual question respectively, the token encoding and fusion modules to encode and exchange the tokenized features from different modalities, and the grounded answer generation modules to predict the final visually grounded answer. 
\vspace{-1em}
\subsection{Tokenization module.}
We tokenize the input question $\boldsymbol{Q}$ and 3D point cloud $\boldsymbol{P}$ into word-level tokens and object-level tokens by using a language tokenizer and a 3D point cloud tokenizer, respectively.

\noindent\textbf{Language Tokenizer.} 
Our language tokenizer is a standard T5 tokenizer~\cite{jmlr_RaffelSRLNMZLL20_T5}, which is widely-used in natural language processing. We also follow T5 to add a task-specific prefix before feeding the question into the T5-tokenizer ( \eg, ``[visual question answering:] is there a chair in this room?''). The outputs of our language tokenizer are the list of IDs of all the words appeared in the question.


\noindent\textbf{3D Point Cloud Tokenizer (\ie, 3D Object Detector).}
The input of the 3D detector is the Point Cloud $\boldsymbol{P} \in \mathbb{R}^{N\times(3+K)}$, which represents the whole scene as $N$ 3D coordinates with $K$ dimensional hidden features, which is a combination of the color, the normal vector, the height, and the 128-dimensional multi-view appearance features for each point in the 3D scene.
For the object detector, we use a variant of VoteNet~\cite{iccv_VoteNet}, in which we do not use the pre-defined anchors but predict the distance between the voted 3D points and the bounding box boundaries, inspired by FCOS~\cite{tian2019fcos}. The outputs of our point cloud tokeinzer are the predicted object proposals for all the objects in the 3D scene.

%

\begin{figure}[t]
\centering
\includegraphics[width=\linewidth]{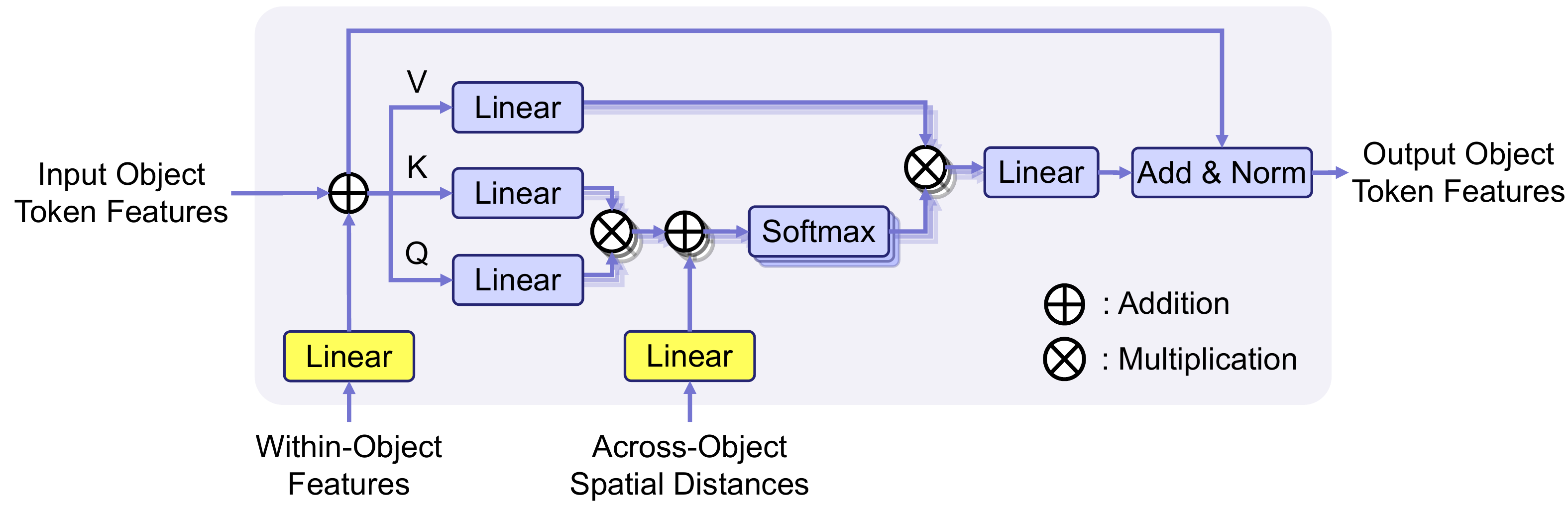}
\vspace{-2em}
\caption{The detailed structure of the enhanced self-attention module for the object branch. We additionally encode within-object attribute features and across-object spatial distances to enhance 3D object token features.}
\label{fig:enhanced_self_attention}
\vspace{-1em}
\end{figure}

\subsection{Token Encoding and Fusion Module}
\label{sec:encoding_fusion_module}

We use $k$ interlaced self-attention and cross-attention layers for token-level feature encoding and fusion. The self-attention layers aim to extract high-level semantic information and encode rich relation information (\ie, word-to-word relations for the question, and object-to-object relations for the point clouds) respectively, while the cross-attention layers focus on how to model complex interactions between different modalities and exchange word-object information.
In our experiments, we empirically set $k=2$.
With the baseline framework in hand, we propose to enhance the 3D VQA performance by approaching to the three objectives more closely via the improved language encoding branch, enhanced point cloud encoding branch, as well as the enriched fusion modules.

Firstly, for the language encoding, we use the encoder of a pre-trained T5 model~\cite{jmlr_RaffelSRLNMZLL20_T5} (\ie, a 12-layer transformer encoder) to extract high-level semantic features and effectively encode the free-form questions. The motivation is that, the self-attention layers in the T5 model could encode rich word-to-word relations. Moreover, the questions in our FE-3DGQA dataset are free-form, and simply encoding the original free-form questions without using any pre-trained language model may be overfited by the questions. Then the language encoding modules not only help predict the labels of the answer-grounded objects that appeared in the question(``AG-In-Q'') and the contextual objects (``Context-Of-AG'') related to the ``AG-In-Q'' objects, but also 
help select the potential answer types.

Secondly, for the 3D object encoding branch, we propose an enhanced self-attention module to extensively encode the within-object features (\ie, the shape, color, texture, and materials) and across-object features (\ie, the complex global \& local relations among different objects) to precisely attend and ground to all of the answer related objects. As shown in Fig.~\ref{fig:enhanced_self_attention}, for the within-object feature encoding, we apply a 2-layer linear projection to the concatenation of the bounding box corners and the inputted $K$ dimensional multi-view RGB features in the corresponding bounding box.
For across-object feature encoding, motivated by the success of relative position bias in the state-of-the-art 2D and 3D methods~\cite{liu2021swin,iccv21_3DVG-Transformer}, we encode the pairwise distances between any two object tokens (\ie, the distance between any two object centers along $x,y$ and $z$ direction) by using a 2-layer linear projection and then add it to the attention map of each self-attention module. The well encoded object features not only provide within-object and across-object features of all the three types of objects, but also help identify the answer-grounded objects not appeared in the question (``AG-NotIn-Q''). Moreover, it also encodes the visual evidence to the correct answers.

Thirdly, the cross-attention-based fusion modules between the language and visual branches focus on how to model complex interactions between different modalities and exchange word-object information. In our model, we use co-attention layers by adding cross-attention layers to both branches, respectively. Specifically, the cross-attention layer in the language branch helps the model to predict the labels of all the three types of answer related objects, because the language branch itself is by no means to predict the answer-grounded object not appeared in the question (``AG-NotIn-Q'') without visual information. It also allows the prediction of the correct answer based on the joint textual and visual features. For the cross-attention layer in the visual branch, the encoded word features in the question provide a guidance on where to attend the feature in the 3D scene for filtering out the unrelated objects features to the potential answer.

\subsection{Grounded Answer Generation Module}
With the encoded and fused features from the question and 3D point clouds, we generate the free-form grounded answers together with the corresponding labels, types, and bounding boxes of the related objects. For answer generation, we use the decoder of a pre-trained T5 model~\cite{jmlr_RaffelSRLNMZLL20_T5} (\ie, a 12-layer transformer decoder) 
with two more parallel linear layers for answer prediction and related objects label and type prediction, respectively. Specifically, for simplicity, we cast the answer generation task as a classification problem similar to the previous 3D VQA methods~\cite{corr2021_ScanQA,corr2021_3D_Question_Answering,corr2021_CLEVR3D}. 
Besides, to improve the explainability and reliability of the predicted answer, we also propose an auxiliary task by predicting the QA-related object classes and the corresponding object types for each class, 
which attempt to link the answer to specific related objects. Besides, the 3D vision branch not only locates and classifies the bounding boxes of all the related objects but also predicts the corresponding object types, such that the predicted answer is completely grounded. 

\subsection{Loss Function}
The loss function of our method is a combination of detection loss $L_\text{detection}$, question answering loss $L_\text{QA}$, the object type classification loss $L_\text{type}$ and an auxiliary classification loss $L_\text{sem-cls}$ to predict the related object semantic classes and the types to each class.
The detection loss $L_\text{detection}$ is almost the same as used in VoteNet~\cite{iccv_VoteNet}, except that the anchor-based bounding box classification loss and the regression loss are replaced with the boundary regression loss~\cite{tian2019fcos}.
We formulate the grounded question answering problem as a classical multi-class classification problem for answers and a multi-label grounding problem for the grounded objects.
We use the focal loss as the question answering loss $L_\text{QA}$ to better handle the long-tailed answers.
The object type classification loss $L_\text{type}$ from the 3D vision branch is a focal loss for the object proposals generated by the detector, where
we regard the object proposals whose IoUs with the ground-truth objects larger than 0.5
as positive samples and other object proposals as negative samples during the training process.
The auxiliary related object semantic label and type classification loss $L_\text{sem-cls}$ of the language branch is a focal loss to predict both object labels and corresponding types.
The final loss function is a combination of these items, \ie, $L_\text{All} = L_\text{detection} + 0.5 L_\text{QA} + 0.5 L_\text{type} + 0.5 L_\text{sem-cls}$.

\begin{table*}[t]
\centering
\caption{
Comparison of the 3D visual question answering (VQA) and grounding results from different methods on the FE-3DGQA dataset. [*]: Results based on our re-implementation. In ours(manual+Ext.), we combine the FE-3DGQA and Ext. FE-3DGQA as the training dataset.}

\begin{tabular}{c|c|ccccc|cccc}
\hline
 & \multicolumn{1}{l|}{} & \multicolumn{5}{c|}{3D VQA Results (Acc)} & \multicolumn{4}{c}{Grounding Results (AP@0.5)} \\ \cline{3-11} 
Methods & \multicolumn{1}{l|}{Input Modality} & number & color & yes/no & other & overall & AG-In-Q & AG-NotIn-Q & Context-Of-AG & mean \\ \hline
QA w/o Scene & - & 29.21 & 32.31 & 55.65 & 17.59 & 37.25 & - & - & - & - \\ \hline
Multiview+MCAN & 2D & 35.08 & 44.42 & 59.73 & 28.99 & 44.50 & - & - & - & - \\
Multiview+T5\&Co-Attn & 2D & 35.07 & 41.25 & 64.44 & 26.81 & 45.16 & - & - & - & - \\ \hline
ScanQA*~\cite{corr2021_ScanQA} & 3D & 40.30 & 35.47 & 62.30 & 29.64 & 44.93 & 20.92 & 4.66 & 7.26 & 10.95 \\
3DQA-TR*~\cite{corr2021_3D_Question_Answering} & 3D & 34.48 & 36.27 & 66.98 & 29.39 & 45.95 & 13.71 & 5.10 & 10.23 & 9.68 \\
Ours & 3D & 34.90 & 40.17 & 68.27 & 31.08 & 47.88 & 28.20 & 11.38 & 19.72 & 19.77 \\ \hline
Ours (manual+Ext.) & 3D & 37.59 & 44.70 & 67.42 & 32.98 & 49.11 & 34.97 & 15.41 & 25.44 & 25.28 \\ \hline
Human & 3D & 85.25 & 70.15 & 90.26 & 50.85 & 74.13 & 81.94 & 53.10 & 52.87 & 62.34 \\ \hline
\end{tabular}
\label{tab:3DGQA_results}
\end{table*}

\begin{table}[t]
\centering
\caption{
Comparison of the results from different methods on validation set of the ScanQA~\cite{corr2021_ScanQA} dataset.}
\begin{tabular}{c|cccc}
\hline
 & EM@1 & EM@10 & Acc@0.25 & Acc@0.5 \\ \hline
\begin{tabular}[c]{@{}c@{}}RandomImage\\ +Oscar~\cite{li2020oscar} (real)\end{tabular} & 19.38 & 46.37 & - & - \\
\begin{tabular}[c]{@{}c@{}}TopDownImage\\ +Oscar~\cite{li2020oscar}\end{tabular} & 17.20 & 43.81 & - & - \\ \hline
ScanRefer+MCAN~\cite{yu2019mcan} & 18.59 & 46.76 & 23.53 & 11.76 \\
ScanQA~\cite{corr2021_ScanQA} & 21.05 & 51.23 & 24.96 & 15.42 \\
Ours & 22.26 & 54.51 & 26.62 & 18.83 \\ \hline
\end{tabular}
\label{tab:ScanQA_results}
\end{table}

\section{Experiments}
To evaluate the newly proposed method, we conduct the experiments based on our newly collected FE-3DGQA dataset (as well as its extended version). 

\subsection{3D Grounded Question Answering Results.}
In Table~\ref{tab:3DGQA_results}, we report the completely grounded 3D visual question answering (VQA) results including not only the QA results, but also the grounding results from three different types of objects. According to different types of answers, we split the validation set into 4 sub-classes: ``number'', ``color'', ``yes/no'', and ``other''.
Since most recent methods and datasets are not open-sourced yet, we compare the results of our method with few state-of-the-art methods~\cite{corr2021_3D_Question_Answering,corr2021_ScanQA} based on our implementation. 
For fair comparison, we also add a grounding head to their methods and also report the grounding results of the three types of objects.
Note that we do not report the results of TransVQA3D~\cite{corr2021_CLEVR3D} since this baseline method requires additional relation graph annotations. 

In general, all the 3D VQA methods outperform the pure QA results without the visual data (\ie, the results from random guess or the language priors), and our method achieves the best overall QA results, as well as the grounding results for all types of objects. 
Specifically, our method achieves around $1.93$\% improvements in the ``overall'' QA results when compared with the recent 3DQA-TR~\cite{corr2021_3D_Question_Answering} method on the validation set. 
For the grounding results, our method achieves remarkable performance gain of $8.82$\% in terms of the mAP$@$0.5 when compared with the  ScanQA~\cite{corr2021_ScanQA} method.
\zlc{We also compare our method with the methods using multiview 2D images with various vision-language fusion methods (\ie, MCAN~\cite{yu2019mcan} or pretrained T5 + co-attention). We found that the method using multiview images performs better for the ``color'' subset, but worse for the ``yes/no'' and ``other'' subsets, which requires 3D scene understanding.}
The results show that our newly proposed method could effectively ground to the answer related objects and thus predict more accurate answers.
For the baseline methods ScanQA~\cite{corr2021_ScanQA} and 3DQA-TR from 3DQA~\cite{corr2021_3D_Question_Answering}, the grounding results are poor, which indicate that their methods could not ground to the answer related objects precisely and thus are not well explainable.
We also conduct the experiments by combining the manually annotated dataset (\ie, FE-3DGQA) and the extended dataset (\ie, Ext. FE-3DGQA) in the training stage to further enhance our performance (see the 5th row in Table.~\ref{tab:3DGQA_results}) for the 3D VQA task. We examine human performance by asking 4 human subjects to answer 400 randomly selected questions and identify the three types of related objects in the 3D scene. The average results are shown in the last row of Table.~\ref{tab:3DGQA_results}. It can be seen that the performance gaps between all the algorithms and human are still large, which allows further exploitation of the 3D GQA task by using our dataset.
As shown in Table.~\ref{tab:ScanQA_results}, we also report the results of our methods on the ScanQA~\cite{corr2021_ScanQA} dataset.
Results show that our method outperforms the baseline methods.


\noindent\textbf{Training Details.}
The model is trained in an end-to-end manner on a machine with a single 32G V100 GPU. For the FE-3DGQA dataset, we train our model for 100 epochs with 15 GPU hours.
When we use the large Extended FE-3DGQA dataset, we train our model for 50 epochs with about 60 GPU hours. We set the learning rate as 1e-3 for the VoteNet~\cite{iccv_VoteNet} detector, and 1e-4 for other transformer-based modules.


\setlength\tabcolsep{3pt}
\begin{table}[t]
\centering
\caption{3D Grounded VQA results of our alternative methods when using different vision modules, language modules, and losses. ``Enhanced SA'' \vs ``Standard SA'' means whether we use the enhanced or the standard self-attention module in the vision branch. ``Bi-GRU'' \vs ``T5'' means whether we use a pretrained T5 module or a randomly initialized Bi-GRU module for language modeling. ``$L_\text{All}$'' and ``$L_\text{detection}+0.5L_\text{QA}$'' mean our overall joint training method and our alternative method without the grounding task.}
\label{tab:analysis}
\begin{tabular}{c|c|c|c|c}
\hline
Vision & Language & Loss & VQA Acc & mAP@0.5 \\ \hline
Standard SA & T5 & $L_\text{All}$ & 47.04 & 19.03 \\
Enhanced SA & Bi-GRU & $L_\text{All}$ & 44.99 & 12.23 \\
Enhanced SA & T5 & $L_\text{detection}+0.5L_\text{QA}$ & 45.99 & - \\ \hline
Enhanced SA & T5 & $L_\text{All}$ & 47.88 & 19.77 \\ \hline
\end{tabular}
\vspace{-1em}
\end{table}

\subsection{Ablation Studies and Analysis}

\begin{figure*}[t]
\centering
\includegraphics[width=0.92\linewidth]{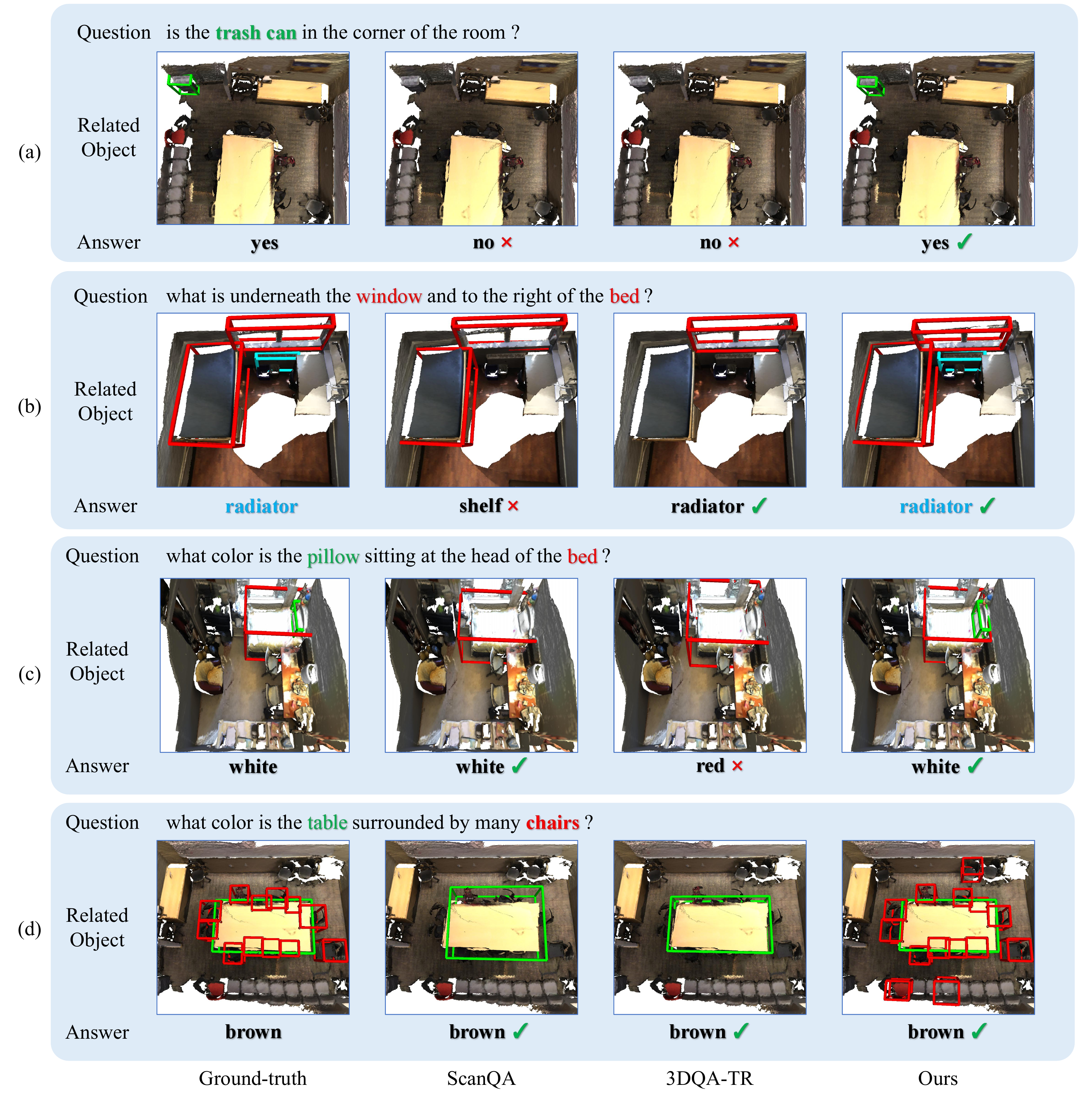}
\caption{Visualization of the results from our method and other baseline methods on the newly proposed FE-3DGQA dataset. The related objects of AG-In-Q, AG-NotIn-Q, and Context-Of-AG are colored in green, cyan, and red, respectively.}
\label{fig:visualization}
\end{figure*}

\begin{figure*}[t]
\centering
\includegraphics[width=0.94\linewidth]{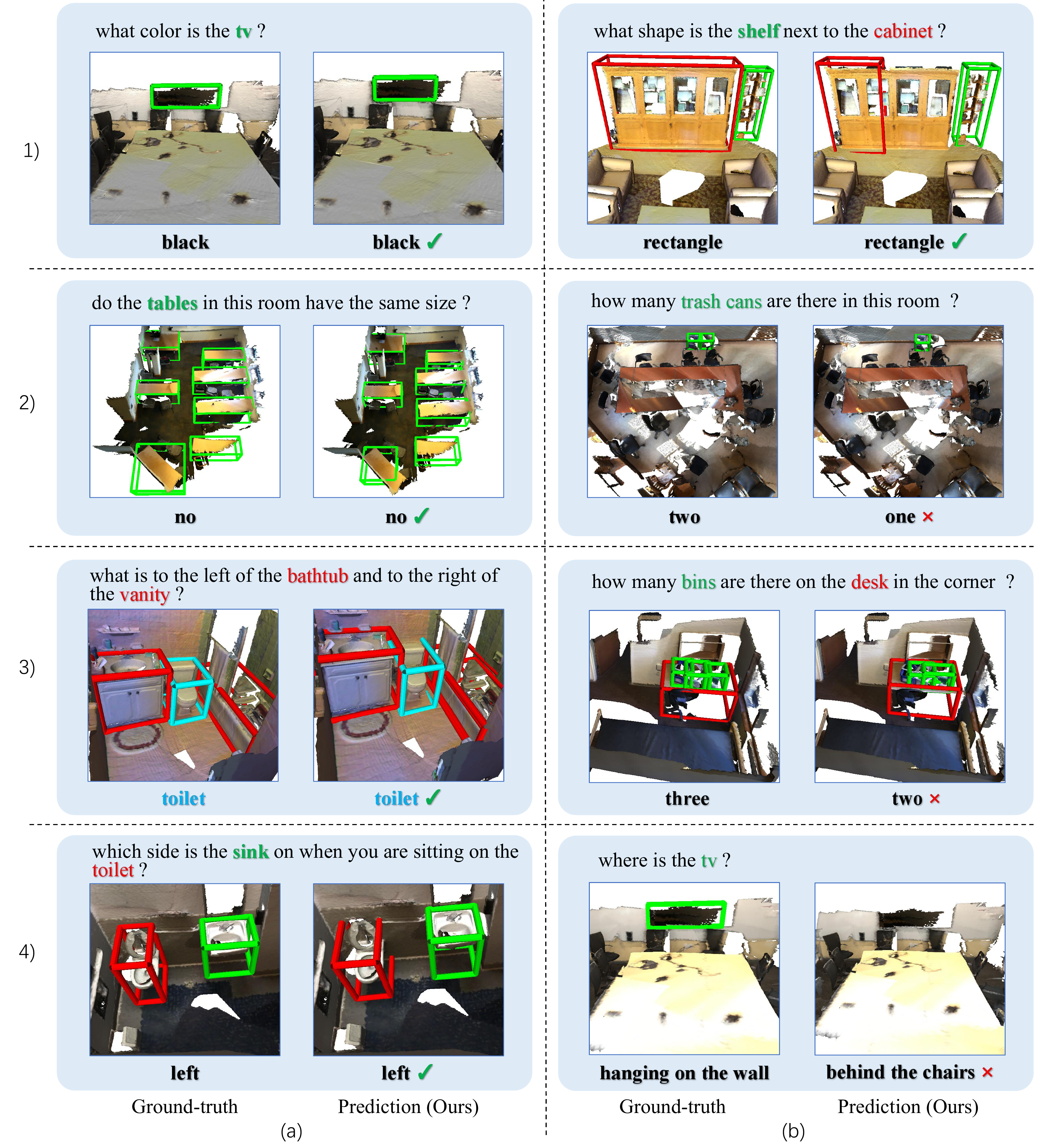}
\caption{More visualization results from our method on the newly collected FE-3DGQA dataset.}
\label{fig:visualization_gt_ours}
\end{figure*}

\begin{table*}[t]
\caption{Grounded question answering results of our newly proposed method when using different training datasets.}
\centering
\small
\begin{tabular}{c|ccccc|cccc}
\hline
 & \multicolumn{5}{c|}{Question Answering Results (Acc)} & \multicolumn{4}{c}{Grounding Results (AP@0.5)} \\ \cline{2-10} 
Training Dataset & number & color & yes/no & other & overall & AG-In-Q & AG-NotIn-Q & Context-Of-AG & mean \\ \hline
manual & 34.90 & 40.17 & 68.27 & 31.08 & 47.88 & 28.20 & 11.38 & 19.72 & 19.77 \\
manual+det. & 35.80 & 40.04 & \textbf{69.08} & 30.10 & 48.12 & 31.50 & 12.12 & 21.61 & 21.74 \\
manual+det.+ground. & 36.60 & 42.15 & 67.35 & \textbf{33.30} & 48.78 & \textbf{35.04} & 14.34 & \textbf{25.44} & 24.94 \\
manual+Ext. & \textbf{37.59} & \textbf{44.70} & 67.42 & 32.98 & \textbf{49.11} & 34.97 & \textbf{15.41} & \textbf{25.44} & \textbf{25.28} \\ \hline
\end{tabular}
\label{tab:different_training_set_result}
\end{table*}

\subsubsection{How does the auxiliary grounding task affect the QA results?}
We argue that the visually grounded results are more reliable and explainable. Here, we compare the QA results between our alternative method
called ``Ours (w/o Grounding)''
without considering the grounding task (\ie, `` $L_\text{detection}+0.5L_\text{QA}$'') and our complete method based on the joint QA and grounding training strategy (\ie, ``$L_\text{All}$''). 
As shown in Lines 3 and 4 of Table~\ref{tab:analysis}, we can see that the grounding results not only provide potential explanation capability to the QA results, but also improve the overall QA results to a certain degree. Moreover, with the additional detection loss $L_\text{detection}$ from the vision branch, our alternative method ``Ours (w/o Grounding)'' (\ie, with 45.99\% Acc.) also outperforms the ``QA w/o Scene'' method (\ie, `` $L_\text{QA}$'', with 37.25\% Acc. as shown in Line 1 of Table~\ref{tab:3DGQA_results}) by 8.74\%, which also verifies that our 3D vision branch and the fusion method can significantly improve the 3D VQA results.
\subsubsection{Effectiveness of the pre-trained language model in the language branch.}
In Lines 2 and 4 of Table~\ref{tab:analysis}, we report the results by using different language modules (\eg, randomly initialized Bi-GRU and pre-trained T5). When compared to the randomly initialized Bi-GRU, our method using the pre-trained T5 language module achieves much better QA results as well as the grounding results. The results verify that our newly collected dataset contains more free-form questions. Without the pre-trained language module, the model is more prone to overfit to the questions and ground to the wrong objects due to insufficient language modeling capability in the Bi-GRU method, leading to poor QA results.
\subsubsection{Ablation study for the enhanced self-attention module in the 3D vision branch.}
As discussed in Sec.~\ref{sec:encoding_fusion_module}, we propose an enhanced self-attention module in the 3D vision branch to extensively encode the within-object features and across-object spatial distances to help the model attend and ground to all the answer related objects. 
In Lines 1 and 4 of Table~\ref{tab:analysis}, we compare the results of our method and the alternative method that uses the standard self-attention modules. We can see that the enhanced self-attention module contributes to both the final QA results and the grounding results.
Specifically, the grounding results of the ``AG-NotIn-Q'' objects (\ie, the answer-grounded objects not appeared in the question) are improved by 2.7\%,
which verifies that a good visual encoding module helps us identify the challenging but essential ``AG-NotIn-Q'' objects. 

\subsection{Visualization of the Grounded VQA Results.} 
Fig.~\ref{fig:visualization} compares the visualization results (\ie, the grounded VQA results) by using the proposed method and two baseline methods 3DQA-TR~\cite{corr2021_3D_Question_Answering} and ScanQA~\cite{corr2021_ScanQA} on the FE-3DGQA dataset. Given the question and the 3D point clouds, we show not only the predicted answer, but also the grounding results of the three types of answer-related objects. The related objects of AG-In-Q, AG-NotIn-Q, and Context-Of-AG are colored in green, cyan, and red, respectively. Fig.~\ref{fig:visualization} (a) shows that the ``AG-In-Q'' object is successfully grounded and classified by our method and the correct answer is also predicted accordingly. While 3DQA-TR~\cite{corr2021_3D_Question_Answering} correctly answers the question in Fig.~\ref{fig:visualization} (b), it fails to detect all the ``Context-Of-AG'' objects and the ``AG-NotIn-Q'' objects, and thus the correct answer is not explainable. By contrast, our method predicts the correct answer together with the completely grounded related objects. In another example (see Fig.~\ref{fig:visualization} (c)), both ScanQA~\cite{corr2021_ScanQA} and our method predict the right answer. However, when compared to our completely grounded results, the ScanQA~\cite{corr2021_ScanQA} method predicts the correct answer even without correctly localizing the ``AG-In-Q'' object and thus the result is not reliable. 
For Fig.~\ref{fig:visualization} (d), while all the methods predict the answer correctly, only our method is completely grounded by not only identifying the ``AG-In-Q'' object, but also most of the ``Context-Of-AG'' objects in the 3D scene.





Fig.~\ref{fig:visualization_gt_ours} shows more qualitative results of the ground-truth annotation and the outputs from our method. As mentioned in Section~\ref{sub:3d_vqa_dataset_collection}, our dataset covers the free-form questions from different aspects, including
1) the local object-oriented questions with attributes and local relations of a specific object, 2) the global context-aware questions to understand the whole scene, 3) the complex relationships among multiple objects, and 4) the direction or location related questions potentially for navigation.

In Fig.~\ref{fig:visualization_gt_ours}, we show more results corresponding to all the four aspects.
Fig.~\ref{fig:visualization_gt_ours} (a) shows the correct and explainable answer predicted by our method. In this example, our method not only predicts the correct answer to the question, but also completely grounds all the objects related to the answer. 
By contrast, as shown in Fig.~\ref{fig:visualization_gt_ours} (b), there are still some failure cases from our method. 
For example, in the 1st row of Fig.~\ref{fig:visualization_gt_ours} (b), though the answer to the question is correct, the grounding results are not complete.
For the 2nd and 3rd row of Fig.~\ref{fig:visualization_gt_ours} (b), the answer to the question is wrong due to the incorrect number of grounded objects in the grounding results.
For the 4th row of Fig.~\ref{fig:visualization_gt_ours} (b), the incorrect answer might be caused by the incorrect grounding results to the related objects. The failure cases verify the challenges of our dataset, and more advanced techniques are required to deal with the challenging cases in our future works.


\subsection{The effect of three data extension strategies.}
\label{sec:extended_dataset_ablation}
In Tab.~\ref{tab:different_training_set_result}, we report the grounded visual question answering results of our newly proposed method based on the test set of our newly collected FE-3DGQA dataset together with different extended training datasets.
In this table, ``manual'' means our original manually annotated FE-3DGQA dataset, ``det.'' represents the transformed 3D detection dataset (\ie, ScanNet), ``ground.'' represents the transformed 3D visual grounding datasets (\ie, ScanRefer and Nr3D), and ``Ext.'' represents the full extended FE-3DGQA dataset, which consists of the transformed detection dataset, the transformed visual grounding dataset, and the transformed masked visual grounding dataset.
Note that as the annotations of the related objects for \zj{the samples from both the transformed 3D grounding datasets and the transformed masked 3D grounding datasets} are incomplete, 
\zj{we remove the corresponding loss functions for ``AG-NotIn-Q'' and ``Context-Of-AG'' (\textit{resp.} ``AG-In-Q'' and ``Context-Of-AG'') in the object type classification loss $L_{type}$ for the vision branch, and the semantic label and type classification loss $L_{sem-cls}$ of the language branch for the samples from the transformed 3D grounding datasets (\textit{resp.} the transformed masked 3D grounding datasets).}

The results show that the overall QA results and the grounding results of our method can be generally improved by extending the manually annotated FE-3DGQA dataset through adding more transformed datasets. The results also verify that our FE-3DGQA dataset can be easily extended by transforming more 3D datasets based on the template-based annotations.

\section{Conclusion}
This work formally defines and addresses the free-form and explainable 3D VQA task. First, we collect a new FE-3DGQA dataset with manually annotated diverse and free-form question-answer pairs and dense and complete grounded bounding box annotations for three different types of answer-related objects. Then, we propose a new 3D VQA framework that effectively predicts the completely visually grounded and explainable answer. The experimental results verify the effectiveness of the collected benchmark dataset for evaluating 3D VQA methods, as well as validate that our proposed framework achieves the state-of-the-art 3D VQA performance by predicting the completely visually grounded answer.

{
\bibliographystyle{IEEEtran}
\bibliography{egbib}

\begin{thebibliography}{10}
\providecommand{\url}[1]{#1}
\csname url@samestyle\endcsname
\providecommand{\newblock}{\relax}
\providecommand{\bibinfo}[2]{#2}
\providecommand{\BIBentrySTDinterwordspacing}{\spaceskip=0pt\relax}
\providecommand{\BIBentryALTinterwordstretchfactor}{4}
\providecommand{\BIBentryALTinterwordspacing}{\spaceskip=\fontdimen2\font plus
\BIBentryALTinterwordstretchfactor\fontdimen3\font minus
  \fontdimen4\font\relax}
\providecommand{\BIBforeignlanguage}[2]{{%
\expandafter\ifx\csname l@#1\endcsname\relax
\typeout{** WARNING: IEEEtran.bst: No hyphenation pattern has been}%
\typeout{** loaded for the language `#1'. Using the pattern for}%
\typeout{** the default language instead.}%
\else
\language=\csname l@#1\endcsname
\fi
#2}}
\providecommand{\BIBdecl}{\relax}
\BIBdecl

\bibitem{kazemzadeh2014referitgame}
S.~Kazemzadeh, V.~Ordonez, M.~Matten, and T.~Berg, ``Referitgame: Referring to
  objects in photographs of natural scenes,'' in \emph{Proceedings of the 2014
  conference on empirical methods in natural language processing (EMNLP)},
  2014, pp. 787--798.

\bibitem{das2017visual_dialog}
A.~Das, S.~Kottur, K.~Gupta, A.~Singh, D.~Yadav, J.~M. Moura, D.~Parikh, and
  D.~Batra, ``Visual dialog,'' in \emph{Proceedings of the IEEE conference on
  computer vision and pattern recognition}, 2017, pp. 326--335.

\bibitem{chen2015microsoft_coco_captions}
X.~Chen, H.~Fang, T.-Y. Lin, R.~Vedantam, S.~Gupta, P.~Doll{\'a}r, and C.~L.
  Zitnick, ``Microsoft coco captions: Data collection and evaluation server,''
  \emph{arXiv preprint arXiv:1504.00325}, 2015.

\bibitem{das2018embodied_question_answering}
A.~Das, S.~Datta, G.~Gkioxari, S.~Lee, D.~Parikh, and D.~Batra, ``Embodied
  question answering,'' in \emph{Proceedings of the IEEE Conference on Computer
  Vision and Pattern Recognition}, 2018, pp. 1--10.

\bibitem{anderson2018vision_and_language_navigation}
P.~Anderson, Q.~Wu, D.~Teney, J.~Bruce, M.~Johnson, N.~S{\"u}nderhauf, I.~Reid,
  S.~Gould, and A.~Van Den~Hengel, ``Vision-and-language navigation:
  Interpreting visually-grounded navigation instructions in real
  environments,'' in \emph{Proceedings of the IEEE conference on computer
  vision and pattern recognition}, 2018, pp. 3674--3683.

\bibitem{iccv_AntolALMBZP15_VQAv1}
S.~Antol, A.~Agrawal, J.~Lu, M.~Mitchell, D.~Batra, C.~L. Zitnick, and
  D.~Parikh, ``{VQA:} visual question answering,'' in \emph{ICCV}, 2015.

\bibitem{cvpr_GoyalKSBP17_VQAv2}
Y.~Goyal, T.~Khot, D.~Summers{-}Stay, D.~Batra, and D.~Parikh, ``Making the {V}
  in {VQA} matter: Elevating the role of image understanding in visual question
  answering,'' in \emph{CVPR}, 2017.

\bibitem{chen2020scanrefer}
D.~Z. Chen, A.~X. Chang, and M.~Nie{\ss}ner, ``{ScanRefer}: {3D} object
  localization in {RGB-D} scans using natural language,'' in \emph{ECCV}, 2020.

\bibitem{chen2021scan2cap}
Z.~Chen, A.~Gholami, M.~Nie{\ss}ner, and A.~X. Chang, ``{Scan2Cap}:
  Context-aware dense captioning in rgb-d scans,'' in \emph{CVPR}, 2021.

\bibitem{iccv21_3D_Visual_Graph_Network_for_Object_Grounding}
M.~Feng, Z.~Li, Q.~Li, L.~Zhang, X.~Zhang, G.~Zhu, H.~Zhang, Y.~Wang, and
  A.~Mian, ``Free-form description guided 3d visual graph network for object
  grounding in point cloud,'' in \emph{ICCV}, 2021.

\bibitem{shih2016where_to_look}
K.~J. Shih, S.~Singh, and D.~Hoiem, ``Where to look: Focus regions for visual
  question answering,'' in \emph{Proceedings of the IEEE conference on computer
  vision and pattern recognition}, 2016, pp. 4613--4621.

\bibitem{corr2021_ScanQA}
D.~Azuma, T.~Miyanishi, S.~Kurita, and M.~Kawanabe, ``Scanqa: 3d question
  answering for spatial scene understanding,'' \emph{arXiv}, vol. 2112.10482,
  2021.

\bibitem{corr2021_3D_Question_Answering}
S.~Ye, D.~Chen, S.~Han, and J.~Liao, ``3d question answering,'' \emph{arXiv},
  vol. 2112.08359, 2021.

\bibitem{corr2021_CLEVR3D}
X.~Yan, Z.~Yuan, Y.~Du, Y.~Liao, Y.~Guo, Z.~Li, and S.~Cui, ``{CLEVR3D:}
  compositional language and elementary visual reasoning for question answering
  in 3d real-world scenes,'' \emph{arXiv}, vol. 2112.11691, 2021.

\bibitem{dai2017scannet}
A.~Dai, A.~X. Chang, M.~Savva, M.~Halber, T.~Funkhouser, and M.~Nie{\ss}ner,
  ``{ScanNet}: Richly-annotated 3d reconstructions of indoor scenes,'' in
  \emph{CVPR}, 2017.

\bibitem{eccv2020_referit3d}
P.~Achlioptas, A.~Abdelreheem, F.~Xia, M.~Elhoseiny, and L.~Guibas,
  ``{ReferIt3D}: Neural listeners for fine-grained 3d object identification in
  real-world scenes,'' in \emph{ECCV}, 2020.

\bibitem{guo2021JointPruning}
J.~Guo, J.~Liu, and D.~Xu, ``{JointPruning}: Pruning networks along multiple
  dimensions for efficient point cloud processing,'' \emph{TCSVT}, 2021.

\bibitem{wangkai2021sequential}
K.~Wang, L.~Sheng, S.~Gu, and D.~Xu, ``Sequential point cloud upsampling by
  exploiting multi-scale temporal dependency,'' \emph{TCSVT}, pp. 4686--4696,
  2021.

\bibitem{liu2021geometrymotion}
J.~Liu and D.~Xu, ``{GeometryMotion-Net}: A strong two-stream baseline for 3d
  action recognition,'' \emph{TCSVT}, pp. 4711--4721, 2021.

\bibitem{csvt_zhao2021transformer3d}
L.~Zhao, J.~Guo, D.~Xu, and L.~Sheng, ``{Transformer3D-Det}: Improving 3d
  object detection by vote refinement,'' \emph{TCSVT}, pp. 4735--4746, 2021.

\bibitem{cvpr_QiSMG17_PointNet}
C.~R. Qi, H.~Su, K.~Mo, and L.~J. Guibas, ``{PointNet}: Deep learning on point
  sets for 3d classification and segmentation,'' in \emph{CVPR}, 2017.

\bibitem{tcsvt_SongZZ22}
Z.~Song, L.~Zhao, and J.~Zhou, ``Learning hybrid semantic affinity for point
  cloud segmentation,'' \emph{TCSVT}, 2022.

\bibitem{tcsvt_WangZLM21}
J.~Wang, H.~Zhu, H.~Liu, and Z.~Ma, ``Lossy point cloud geometry compression
  via end-to-end learning,'' \emph{TCSVT}, 2021.

\bibitem{tcsvt_MekuriaBC17}
R.~Mekuria, K.~Blom, and P.~C{\'{e}}sar, ``Design, implementation, and
  evaluation of a point cloud codec for tele-immersive video,'' \emph{TCSVT},
  2017.

\bibitem{tcsvt_SongSGWL21}
F.~Song, Y.~Shao, W.~Gao, H.~Wang, and T.~Li, ``Layer-wise geometry aggregation
  framework for lossless lidar point cloud compression,'' \emph{TCSVT}, 2021.

\bibitem{iccv21_3DVG-Transformer}
L.~Zhao, D.~Cai, L.~Sheng, and D.~Xu, ``3dvg-transformer: Relation modeling for
  visual grounding on point clouds,'' in \emph{ICCV}, 2021.

\bibitem{arXiv_2021d3net}
D.~Z. Chen, Q.~Wu, M.~Nie{\ss}ner, and A.~X. Chang, ``D3net: {A}
  speaker-listener architecture for semi-supervised dense captioning and visual
  grounding in {RGB-D} scans,'' \emph{arXiv}, vol. 2112.01551, 2021.

\bibitem{cvpr_Wald2020_3DSSG}
J.~Wald, H.~Dhamo, N.~Navab, and F.~Tombari, ``{Learning 3D Semantic Scene
  Graphs from 3D Indoor Reconstructions},'' in \emph{CVPR}, 2020.

\bibitem{iccv_Wald2019RIO_3RScan}
J.~Wald, A.~Avetisyan, F.~T. Nassir~Navab, and M.~Niessner, ``{RIO:} 3d object
  instance re-localization in changing indoor environments,'' in \emph{ICCV},
  2019.

\bibitem{yuan2021instancerefer}
Z.~Yuan, X.~Yan, Y.~Liao, R.~Zhang, Z.~Li, and S.~Cui, ``{InstanceRefer}:
  Cooperative holistic understanding for visual grounding on point clouds
  through instance multi-level contextual referring,'' \emph{ICCV}, 2021.

\bibitem{aaai_huang2021_TGNN}
P.-H. Huang, H.-H. Lee, H.-T. Chen, and T.-L. Liu, ``Text-guided graph neural
  networks for referring 3d instance segmentation,'' in \emph{AAAI}, 2021.

\bibitem{PMLR21_corl_RohDFF21_LanguageRefer}
J.~Roh, K.~Desingh, A.~Farhadi, and D.~Fox, ``Languagerefer: Spatial-language
  model for 3d visual grounding,'' in \emph{PMLR}, 2021.

\bibitem{iccv21_sat}
Z.~Yang, S.~Zhang, L.~Wang, and J.~Luo, ``{SAT:} 2d semantics assisted training
  for 3d visual grounding,'' \emph{ICCV}, 2021.

\bibitem{cvpr21_sunrefer}
H.~Liu, A.~Lin, X.~Han, L.~Yang, Y.~Yu, and S.~Cui, ``Refer-it-in-rgbd: {A}
  bottom-up approach for 3d visual grounding in {RGBD} images,'' in
  \emph{CVPR}, 2021.

\bibitem{he2021_transrefer3d}
D.~He, Y.~Zhao, J.~Luo, T.~Hui, S.~Huang, A.~Zhang, and S.~Liu, ``Transrefer3d:
  Entity-and-relation aware transformer for fine-grained 3d visual grounding,''
  in \emph{ACM MM}, 2021.

\bibitem{wacv_AbdelreheemUSYC22_3DRefTransformer}
A.~Abdelreheem, U.~Upadhyay, I.~Skorokhodov, R.~A. Yahya, J.~Chen, and
  M.~Elhoseiny, ``3dreftransformer: Fine-grained object identification in
  real-world scenes using natural language,'' in \emph{WACV}, 2022.

\bibitem{nips_MalinowskiF14_DAQUAR}
M.~Malinowski and M.~Fritz, ``A multi-world approach to question answering
  about real-world scenes based on uncertain input,'' in \emph{NIPS}, 2014.

\bibitem{zhou2015simple_baseline}
B.~Zhou, Y.~Tian, S.~Sukhbaatar, A.~Szlam, and R.~Fergus, ``Simple baseline for
  visual question answering,'' \emph{arXiv preprint arXiv:1512.02167}, 2015.

\bibitem{teney2018tips_and_tricks}
D.~Teney, P.~Anderson, X.~He, and A.~Van Den~Hengel, ``Tips and tricks for
  visual question answering: Learnings from the 2017 challenge,'' in
  \emph{Proceedings of the IEEE conference on computer vision and pattern
  recognition}, 2018, pp. 4223--4232.

\bibitem{lu2016hierarchical}
J.~Lu, J.~Yang, D.~Batra, and D.~Parikh, ``Hierarchical question-image
  co-attention for visual question answering,'' \emph{Advances in neural
  information processing systems}, vol.~29, 2016.

\bibitem{li2020oscar}
X.~Li, X.~Yin, C.~Li, P.~Zhang, X.~Hu, L.~Zhang, L.~Wang, H.~Hu, L.~Dong,
  F.~Wei \emph{et~al.}, ``Oscar: Object-semantics aligned pre-training for
  vision-language tasks,'' in \emph{European Conference on Computer
  Vision}.\hskip 1em plus 0.5em minus 0.4em\relax Springer, 2020, pp. 121--137.

\bibitem{lu2019vilbert}
J.~Lu, D.~Batra, D.~Parikh, and S.~Lee, ``Vilbert: Pretraining task-agnostic
  visiolinguistic representations for vision-and-language tasks,''
  \emph{Advances in neural information processing systems}, vol.~32, 2019.

\bibitem{jmlr_RaffelSRLNMZLL20_T5}
C.~Raffel, N.~Shazeer, A.~Roberts, K.~Lee, S.~Narang, M.~Matena, Y.~Zhou,
  W.~Li, and P.~J. Liu, ``Exploring the limits of transfer learning with a
  unified text-to-text transformer,'' \emph{J. Mach. Learn. Res.}, 2020.

\bibitem{iccv_VoteNet}
C.~R. Qi, O.~Litany, K.~He, and L.~J. Guibas, ``Deep hough voting for 3d object
  detection in point clouds,'' in \emph{ICCV}, 2019.

\bibitem{tian2019fcos}
Z.~Tian, C.~Shen, H.~Chen, and T.~He, ``Fcos: Fully convolutional one-stage
  object detection,'' in \emph{ICCV}, 2019, pp. 9627--9636.

\bibitem{liu2021swin}
Z.~Liu, Y.~Lin, Y.~Cao, H.~Hu, Y.~Wei, Z.~Zhang, S.~Lin, and B.~Guo, ``Swin
  transformer: Hierarchical vision transformer using shifted windows,'' in
  \emph{ICCV}, 2021.

\bibitem{yu2019mcan}
Z.~Yu, J.~Yu, Y.~Cui, D.~Tao, and Q.~Tian, ``Deep modular co-attention networks
  for visual question answering,'' in \emph{Proceedings of the IEEE/CVF
  conference on computer vision and pattern recognition}, 2019, pp. 6281--6290.

\end{thebibliography}
}


\end{document}